\documentclass[11pt,a4paper]{article}
\usepackage[hyperref,acceptedWithA]{tacl2018v2} % use ``nohyperref'' to disable  hyperref
\usepackage{times,latexsym}
\usepackage{url}
\usepackage[T1]{fontenc}

\usepackage[english]{babel}
\usepackage{xspace,mfirstuc,tabulary}

\usepackage{paralist} 
\usepackage{graphicx} 
\usepackage{multirow} 
\usepackage{enumitem}
\usepackage{linguex}
\title{\emph{Tabula} nearly \emph{rasa:} Probing the linguistic knowledge of character-level neural language models trained on unsegmented text}

\author{Michael Hahn\thanks{Work partially done while interning at Facebook AI Research.} \\ Stanford University \\ \url{mhahn2@stanford.edu} \And Marco Baroni \\ Facebook AI Research \\ UPF Linguistics Department \\ Catalan Institution for Research\\and Advanced Studies \\ \url{mbaroni@gmail.com}}

\date{}

\hypersetup{draft}

\begin{document}
\maketitle
\begin{abstract}
  Recurrent neural networks (RNNs) have reached striking performance in
  many natural language processing tasks. This has renewed interest in
  whether these generic sequence processing devices are inducing
  genuine linguistic knowledge. Nearly all current analytical studies,
  however, initialize the RNNs with a vocabulary of known words, and
  feed them tokenized input during training. We present a
  multi-lingual study of the linguistic knowledge encoded in RNNs
  trained as character-level language models, on input data with word
  boundaries removed. These networks face a tougher and more
  cognitively realistic task, having to discover any useful
  linguistic unit from scratch based on input statistics. The results
  show that our ``near \emph{tabula rasa}'' RNNs are mostly able to
  solve morphological, syntactic and semantic tasks that intuitively
  presuppose word-level knowledge, and indeed they learned, to some extent, to track
   word boundaries. Our study opens the door to speculations
  about the necessity of an explicit, rigid word lexicon in language learning and
  usage.
\end{abstract}

\section{Introduction}
\label{sec:introduction}

Recurrent neural networks \cite[RNNs,][]{Elman:1990}, in particular
in their Long-Short-Term-Memory variant
\cite[LSTMs,][]{Hochreiter:Schmidhuber:1997}, are widely used in natural language processing. RNNs, often
pre-trained on the simple \emph{language modeling} objective of
predicting the next symbol in natural text, are often a crucial
component of state-of-the-art architectures for machine
translation, natural language inference and text categorization
\cite{Goldberg:2017}.

RNNs are very general devices for sequence processing, hardly assuming
any prior linguistic knowledge. Moreover, the simple prediction task
they are trained on in language modeling is well-attuned to the core
role prediction plays in cognition
\cite[e.g.,][]{Bar:2007,Clark:2016}. RNNs have thus long attracted
researchers interested in language acquisition and processing. Their
recent successes in large-scale tasks has rekindled
this interest \cite[e.g.,][]{Frank:etal:2013,Lau:etal:2017,Kirov:Cotterell:2018,Linzen:etal:2018,McCoy:etal:2018,Pater:2018}.

The standard pre-processing pipeline of modern RNNs assumes that the
input has been tokenized into word units that are pre-stored in the
RNN vocabulary \cite{Goldberg:2017}. This is a reasonable practical
approach, but it makes simulations less interesting from a linguistic
point of view. First, discovering words (or other primitive
constituents of linguistic structure) is one of the major challenges a
learner faces, and by pre-encoding them in the RNN we are facilitating
its task in an unnatural way (not even the staunchest nativists would
take specific word dictionaries to be part of our genetic
code). Second, assuming a unique tokenization into a finite number of
discrete word units is in any case problematic. The very notion of
what counts as a word in languages with a rich morphology is far from
clear \cite[e.g.,][]{Dixon:Aikhenvald:2002,Bickel:Zuniga:2017}, and,
universally, lexical knowledge is probably organized into a
not-necessarily-consistent hierarchy of units at different levels:
morphemes, words, compounds, constructions,
etc.~\cite[e.g.,][]{Goldberg:2005}. Indeed, it has been suggested that
the notion of word cannot even be meaningfully defined
cross-linguistically \cite{Haspelmath:2011}.

Motivated by these considerations, we study here RNNs that are trained
without any notion of word in their input or in their
architecture. We train our RNNs as \emph{character-level neural
  language models}
\cite[CNLMs,][]{Mikolov:etal:2011,Sutskever:etal:2011,DBLP:journals/corr/Graves13}
by removing whitespace from their input, so that, like children
learning a language, they don't have access to explicit cues to
wordhood.\footnote{We do not erase punctuation marks, reasoning that
  they have a similar function to prosodic cues in spoken language.}
This setup is almost as \emph{tabula rasa} as it gets. By using
unsegmented orthographic input (and assuming that, in the alphabetic
writing systems we work with, there is a reasonable correspondence
between letters and phonetic segments), we are only postulating that
the learner figured out how to map the continuous speech stream to a
sequence of phonological units, an ability children already possess
few months after birth \cite[e.g.,][]{Maye:etal:2002,Kuhl:2004}. We believe that focusing on language modeling of an unsegmented phoneme sequence, abstracting away from other complexities of a fully realistic child language acquisition setup, is particularly instructive, in order to study which linguistic structures naturally emerge.

We evaluate our character-level networks on a bank of linguistic tests
in German, Italian and English. We focus on these languages due to
resource availability and ease of benchmark construction. Also, well-studied synthetic languages with a clear,
orthographically-driven notion of word might be a better starting point to
test non-word-centric models, compared to agglutinative or
polysynthetic languages, where the very notion of what counts as a
word is problematic. % While one of
% our ultimate goals is precisely to study how word-less models process
% languages whose grammatical system is less clearly word-based,
% starting with languages in whose analysis the orthographic word has
% traditionally played a central role is a reasonable ``sanity check''.
  
Our tasks require models to develop the latent
ability to parse characters into word-like items associated to
morphological, syntactic and broadly semantic features. The RNNs
pass most of the tests, suggesting that they are in some way able to
construct and manipulate the right lexical objects. In a final experiment,
we look more directly into \emph{how} the models are handling
word-like units. We find, confirming an earlier observation by
\newcite{Kementchedjhieva:Lopez:2018}, that the RNNs specialized some
cells to the task of detecting word boundaries (or, more generally,
salient linguistic boundaries, in a sense to be further discussed
below). Taken together, our results suggest that character-level RNNs
capture forms of linguistic knowledge that are traditionally thought to be
word-based, without being exposed to an explicit segmentation of their
 input and, more importantly, without possessing an explicit word
lexicon. We will discuss the implications of these findings in the
conclusion.\footnote{Our input data,
  test sets and pre-trained models are available at \url{https://github.com/m-hahn/tabula-rasa-rnns}.}

% probe them with phonological, lexical,
% morphological, syntactic and semantic tests in English, German and
% Italian. Our results show that near-\emph{tabula-rasa} CNLMs acquire
% an impressive spectrum of linguistic knowledge at various levels.
% This in turn suggests that, given abundant input (large Wikipedia
% dumps), a learning device whose only prior architectural bias consists
% in the LSTM memory cell implicitly acquires a variety of linguistic
% rules that one would intuitively expect to require much more prior
% knowledge.

\section{Related work}
\label{sec:related}

\paragraph{On the primacy of words} Several linguistic studies suggest
that words, at least as delimited by whitespace in some writing
systems, are neither necessary nor sufficient units of linguistic
analysis. \newcite{Haspelmath:2011} claims that there is no
cross-linguistically valid definition of the notion of word \cite[see also][who
address specifically the notion of prosodic
word]{Schiering:etal:2010}. Others have stressed the difficulty of
characterizing words in polysynthetic languages
\cite{Bickel:Zuniga:2017}. Children are only rarely exposed to words
in isolation during learning
\cite{Tomasello:2003},\footnote{Single-word utterances are not
  uncommon in child-directed language, but they are still rather the
  exception than the rule, and many important words, such as
  determiners, never occur in isolation
  \cite{Christiansen:etal:2005}.} and it is likely that the units that
adult speakers end up storing in their lexicon are of variable size,
both smaller and larger than conventional words
\cite[e.g.,][]{Jackendoff:2002,Goldberg:2005}. From a more applied
perspective, \newcite{Schuetze:2017} recently defended tokenization-free
approaches to NLP, proposing a general non-symbolic approach to text
representation. %
%Our
% study shows that a powerful sequence learning device, such as an RNN
% with LSTM cells, can learn, from naturally occurring language data, to
% capture several linguistic phenomena that appear to be word-mediated
% without overt word-boundary information and lacking an internal data
% structure for a word vocabulary. The model, moreover, develops during
% learning units that track word-like boundaries, despite its lack of an
% explicit lexicon.
We hope our results will contribute to the theoretical
debate on word primacy, suggesting, through computational simulations, that
word priors are not crucial to language learning and processing.

\paragraph{Character-based neural language models} received attention in the last
decade because of their greater generality  compared to word-level models. % , and because, intuitively, they should be able to
% use cues, such as morphological information, that word-based models
% miss by design.
Early studies
\cite{Mikolov:etal:2011,Sutskever:etal:2011,DBLP:journals/corr/Graves13}
established that CNLMs might not be as good at language modeling as
their word-based counterparts, but lag only slightly behind. This is
particularly encouraging in light of the fact that character-level
sentence prediction involves a much larger search space than
prediction at the word level, as a character-level model must make a
prediction after each character, rather than after each
word. \newcite{Sutskever:etal:2011} and
\newcite{DBLP:journals/corr/Graves13} ran qualitative analyses showing
that CNLMs capture some basic linguistic properties of their
input. The latter, who used LSTM cells, also showed, qualitatively,
that CNLMs are sensitive to hierarchical structure. In particular,
they balance parentheses correctly when generating text. %
% Our aim here is to understand to what extent CNLMs trained on
% unsegmented input learn various linguistic constructs. % This differs
% from

Most recent work in the area has focused on \emph{character-aware}
architectures combining character- and word-level information to
develop state-of-the-art language models that are also effective in
morphologically rich languages
\citep[e.g.,][]{Bojanowski:etal:2016,Kim:etal:2016,Gerz:etal:2018}. For
example, Kim and colleagues perform prediction at the word level, but
use a character-based convolutional network to generate word
representations. Other work focuses on splitting words into morphemes,
using character-level RNNs and an explicit segmentation objective
\cite[e.g.,][]{Kann:etal:2016}. These latter lines of work are only
distantly related to our interest in probing what a purely
character-level network trained on running text has implicitly learned
about linguistic structure. There is also extensive work on
segmentation of the linguistic signal that does not rely on neural
methods, and is not directly relevant here, \cite[e.g.,][and references there]{Brent:Cartwright:1996,goldwater-bayesian-2009,Kamper:etal:2016}.
% about morphemes and other units through generic language
% modeling.% Moreover, these networks are not exposed to constituents
% larger than words.

\paragraph{Probing linguistic knowledge of neural language models} is
currently a popular research topic
\cite{Li:etal:2016,Linzen:etal:2016,Shi:etal:2016,Adi:etal:2017,Belinkov:etal:2017,Kadar:etal:2017,Hupkes:etal:2017,Conneau:etal:2018,Ettinger:etal:2018,Linzen:etal:2018}. Among
studies focusing on character-level models, 
\newcite{Elman:1990} already reported a proof-of-concept experiment on
implicit learning of word
segmentation. \newcite{Christiansen:etal:1998} trained a RNN on
phoneme-level language modeling of transcribed child-directed speech
with tokens marking utterance boundaries, and found that the network
learned to segment the input by predicting the utterance boundary
symbol also at word edges. More recently, \newcite{Sennrich:2017}
explored the grammatical properties of character- and
subword-unit-level models that are used as components of a machine
translation system. He concluded that current character-based decoders
generalize better to unseen words, but capture less grammatical
knowledge than subword units. Still, his character-based systems
lagged only marginally behind the subword architectures on grammatical
tasks such as handling agreement and
negation. \newcite{DBLP:journals/corr/RadfordJS17} focused on CNLMs
deployed in the domain of sentiment analysis, where they found the
network to specialize a unit for sentiment tracking. We will discuss
below how our CNLMs also show single-unit specialization, but for boundary
tracking. \newcite{Godin:etal:2018} investigated the rules implicitly
used by supervised character-aware neural morphological segmentation
methods, finding linguistically sensible
patterns. \newcite{Alishahi:etal:2017} probed the linguistic knowledge
induced by a neural network that receives unsegmented acoustic
input. Focusing on phonology, they found that the lower layers of the model process finer-grained information,
whereas higher layers are sensitive to more abstract
patterns. %Their model is not directly comparable to ours, since it
% uses a considerably more complex architecture and it is trained on
% multimodal data.
\newcite{Kementchedjhieva:Lopez:2018} recently probed the linguistic
knowledge of an English CNLM trained with whitespace in the
input. Their results are aligned with ours. The model is sensitive to
lexical and morphological structure, and it captures morphosyntactic
categories as well as constraints on possible morpheme
combinations. Intriguingly, the model tracks word/morpheme boundaries
through a single specialized unit, suggesting that such boundaries are
salient (at least when marked by whitespace, as in their experiments)
and informative enough that it is worthwhile for the network to devote a
special mechanism to process them. We replicated this finding for our
networks trained on whitespace-free text, as discussed in Section
\ref{sec:segmentation} below, where we discuss it in the context of our
other results.

%We were not able to find a similar sparse
%encoding of morpheme- or word-boundaries in our network (the results
%we will report in Section \ref{sec:segmentation} suggest that our
%model is also tracking boundaries in its hidden state, but through a
%distributed code). The most obvious difference between our experiment
%and that of Kementchedjhieva and Lopez is that they keep whitespace in
%the input training.  However, they observe that their
%morpheme-tracking unit is not simply predicting white space,
%suggesting that the encoding difference between our model and theirs
%cannot be (entirely) explained away by the difference in training
%input. As we do not attempt here to characterize the inner dynamics of
%our network, we leave a systematic comparison to future work.

% They do not explore syntactic or semantic
% knowledge, and they limit their study to English. Moreover, they
% trained their models on input with whitespace, thus providing the
% model with a major (and cognitively artificial) cue to word
% boundaries.

\section{Experimental setup}
\label{sec:setup}

We extracted plain text from full English, German and Italian
Wikipedia dumps with
WikiExtractor.\footnote{\url{https://github.com/attardi/wikiextractor}}
We randomly selected test and validation sections consisting of 50,000
paragraphs each, and used the remainder for training. The training
sets contained 16M (German), 9M (Italian), and 41M (English)
paragraphs, corresponding to 819M, 463M and 2,333M words,
respectively. Paragraph order was shuffled for training, without
attempting to split by sentences. All characters were lower-cased.
For benchmark construction and word-based model training, we tokenized
and tagged the corpora with
TreeTagger~\citep{schmid1999improvements}.\footnote{\url{http://www.cis.uni-muenchen.de/~schmid/tools/TreeTagger/}}
We used as vocabularies the most frequent characters from each corpus,
setting thresholds so as to ensure that all characters representing
phonemes were included, resulting in vocabularies of sizes 60
(English), 73 (German), and 59 (Italian).  We further constructed
\emph{word-level neural language models} (WordNLMs); their vocabulary
included the most frequent 50,000 words per corpus.

We trained RNN and LSTM CNLMs; we will refer to them simply as
\emph{RNN} and \emph{LSTM}, respectively. The ``vanilla'' RNN will
serve as a baseline to ascertain if/when the longer-range
information-tracking abilities afforded to the LSTM by its gating
mechanisms are necessary. Our WordNLMs are always LSTMs.  For each
model/language, we applied random hyperparameter search.  We
terminated training after 72 hours.\footnote{This was due to resource
  availability. The reasonable language-modeling results in Table
  \ref{tab:lm-results} suggest that no model is seriously underfit,
  but the weaker overall RNN results in particular should be
  interpreted in the light of the following qualification: models are
  compared \emph{given equal amount of training, but possibly at
    different convergence stages}.} None of the models had overfitted,
as measured by performance on the validation
set.\footnote{Hyperparameter details are in the appendix. Chosen architectures
  (layers/embedding size/hidden size): LSTM: En.~3/200/1024,
  Ge.~2/100/1024, It.~2/200/1024; RNN: En.~2/200/2048, Ge.~2/50/2048,
  It.~same; WordNLM; En.~2/1024/1024, Ge.~2/200/1024, It.~same.}

Language modeling performance on the test partitions is shown in
Table \ref{tab:lm-results}. Recall that we removed whitespace, which
is both easy to predict, and aids prediction of other
characters. Consequently, the fact that our character-level models are
below the state of the art is expected.\footnote{Training our models
  with whitespace, without further hyperparameter tuning, resulted in
  BPCs of 1.32 (English), 1.28 (German), and 1.24 (Italian).}
For example, the best model of \newcite{merity2018analysis} achieved
1.23 English BPC on a Wikipedia-derived dataset. % (Hutter 2018). %, and 1.175 on a version of PTBenglish 0.85 german 0.9, italian 0.82
On EuroParl data, \newcite{cotterell2018all} report 0.85 for English,
0.90 for German, and 0.82 for Italian. Still, our English BPC is
comparable to that reported by \newcite{DBLP:journals/corr/Graves13} for his static
character-level LSTM trained on space-delimited Wikipedia data,
suggesting that we are achieving reasonable performance. %
%\footnote{Training our models on text with whitespace, without further hyperparameter tuning to adjust to that setting, resulted in cross-entropies of 0.91, 1.32 BPC (English), 0.89, 1.28 BPC (German), and 0.86, 1.24 BPC (Italian).}
The perplexity of the word-level model might not be comparable to
that of highly-optimized state-of-the-art architectures, but it is at the
expected level for a well-tuned vanilla LSTM language model. For
example, \newcite{Gulordava:etal:2018} report 51.9 and 44.9 perplexities respectively in English and Italian for
their best LSTMs trained on Wikipedia data with same vocabulary
size as ours.
%=======
%Performance on the test partitions is shown in Table~\ref{tab:lm-results}.
%Direct comparison with the state-of-the-art in character-based language modeling is hindered by the fact that we train on text without whitespace.
%The best models of \cite{merity2018analysis} achieved 1.232 BPC on enwiki8 \cite{hutter2018}, a dataset also derived from English Wikipedia. % (Hutter 2018). %, and 1.175 on a version of PTBenglish 0.85 german 0.9, italian 0.82
%On Europarl data, \cite{cotterell2018all} report 0.85 for English, 0.9 for German, and 0.82 for Italian. 
%Our BPC values are higher, but this is expected given that we do not provide whitespace to the model: Whitespace is both relatively easy to predict, and it makes predicting other characters easier.\footnote{Refitting our models to data with whitespace, without retuning hyperparameters, yields ....}
%>>>>>>> 86b9fd533dc18bab83a010158126d7366aae3681

\begin{table}[t]
  \begin{small}
  \begin{center}
    \begin{tabular}{l|c|c|c}
      \multicolumn{1}{c|}{}&\emph{LSTM}&\emph{RNN}&\emph{WordNLM}\\
      \hline
	    English & 1.62 & 2.08 & 48.99  \\
	    German &  1.51 & 1.83 & 37.96   \\
	    Italian & 1.47 & 1.97 & 42.02  \\
    \end{tabular}
  \end{center}
  \end{small}
  \caption{\label{tab:lm-results} Performance of language models. For CNLMs, we report bits-per-character (BPC). For WordNLMs, we report perplexity.}
\end{table}

%\begin{table}[t]
%  \begin{center}
%    \begin{tabular}{l|l|l|l|l}
%      \multicolumn{1}{c}{}&\emph{LSTM}&\emph{RNN}&\emph{Word LSTM}\\
%      \hline
%	    English & 1.12 / 1.62 & 1.44 / 2.08 & 3.89 / 48.99  \\
%	    German &  1.05 / 1.51 & 1.27 / 1.83 & 3.63 / 37.96   \\
%	    Italian & 1.02 / 1.47 & 1.37 / 1.97 & 3.85 / 42.02  \\
%    \end{tabular}
%  \end{center}
%  \caption{\label{tab:lm-results} Performance of language models. For CNLMs, we report cross-entropy and bits-per-character (BPC). For word-based models, we report cross-entropy and perplexity.}
%\end{table}

\section{Experiments}
\label{sec:experiments}

%\textbf{Todo: add brief outline.}

%\textbf{We could rename the word-based model WNLM, if this can save space in tables.}

% We will probe the linguistic knowledge of the CNLM on different levels of linguistic abstraction, including morphology, word segmentation, morphology, syntax, and semantics.

%\input{phonology}

\subsection{Discovering morphological categories}
\label{sec:categories}

Words belong to part-of-speech categories, such as nouns and
verbs. Moreover, they typically carry inflectional features such as
number. We start by probing whether CNLMs capture such
properties. We use here the popular method of ``diagnostic
classifiers'' \cite{Hupkes:etal:2017}. That is, we treat the hidden
activations produced by a CNLM whose weights were fixed after language
model training as input features for a shallow (logistic) classifier
of the property of interest (e.g., plural vs.~singular). If the
classifier is successful, this means that the representations provided
by the model are encoding the relevant information.  The classifier is
deliberately shallow and trained on a small set of examples, as we
want to test whether the properties of interest are robustly encoded
in the representations produced by the CNLMs, and amenable to a simple
linear readout \cite{Fusi:etal:2016}. In our case, we want to probe
word-level properties in models trained at the character level. To do
this, we let the model read each target word character-by-character,
and we treat the state of its hidden layer after processing the last
character in the word as the model's implicit representation of the
word, on which we train the diagnostic classifier. %
% Besides being sensitive, to some extent, to word boundaries, does
% the CNLM also store linguistic properties of words, such as their part
% of speech and number?
The experiments focus on German and Italian, as it's harder to design
reliable test sets for the impoverished English morphological system.

\paragraph{Word classes (nouns vs.~verbs)}

For both German and Italian, we sampled 500 verbs and 500 nouns from
the Wikipedia training sets, requiring that they are unambiguously
tagged in the corpus by TreeTagger. Verbal and nominal forms are often
cued by suffixes. We removed this confound by selecting examples with
the same ending across the two categories (\emph{-en} in German:
\emph{Westen} `west',\footnote{German nouns are capitalized; this cue
  is unavailable to the CNLM as we lower-case the input.} %
\emph{stehen} `to stand'; and \emph{-re} in Italian: \emph{autore}
`author', \emph{dire} `to say'). We randomly selected 20 training examples
(10 nouns and 10 verbs), and tested on the remaining items.  We
repeated the experiment 100 times to account for random train-test
split variation.
%  We recorded the final
% hidden state of a pre-trained CNLM after reading a word, without
% context, and trained a logistic noun-verb classifier on these
% representations.

While we controlled for suffixes as described above, it could still be
the case that other substrings reliably cue verbs or nouns. We thus
considered a baseline trained on word-internal information only,
namely a character-level LSTM autoencoder trained on the Wikipedia datasets to reconstruct words 
in isolation.\footnote{The autoencoder is implemented as a standard LSTM sequence-to-sequence model~\citep{sutskever2014sequence}. For each language, autoencoder hyperparameters were chosen using random search, as for the language models; details are in supplementary material to be made available upon publication. For both German and Italian models, the following parameters were chosen: 2 layers, 100 embedding dimensions, 1024 hidden dimensions.}
The hidden state of the LSTM autoencoder should capture
discriminating orthographic features, but, by design, will have
no access to broader contexts.  We further considered word embeddings
from the output layer of the WordNLM. Unlike CNLMs, the WordNLM cannot
make educated guesses about words that are not in its training
vocabulary. These OOV words are by construction less frequent, and
thus likely to be in general more difficult. To get a sense of both
``best-case-scenario'' and more realistic WordNLM performance, we
report its accuracy both excluding and including OOV items
(WordNLM$_{\textit{subs.}}$ and WordNLM in Table
\ref{tab:pos-results}, respectively). In the latter case, we let the
model make a random guess for OOV items.  The percentage of OOV items
over the entire dataset, balanced for nouns and verbs, was 92.3\% for
German and 69.4\% for Italian.
Note that none of the words were OOV for the CNLM, as they all were taken from the Wikipedia training set.

Results are in Table~\ref{tab:pos-results}.  All language models
outperform the autoencoders, showing that they learned categories
based on broader distributional evidence, not just typical strings
cuing nouns and verbs. Moreover, the LSTM CNLM outperforms the RNN,
probably because it can track broader contexts. Not surprisingly, the
word-based model fares better on in-vocabulary words, but the gap,
especially in Italian, is rather narrow, and there is a strong
negative impact of OOV
words (as expected, given that WordNLM is at random on them). % Figure~\ref{fig:pos-induction} shows how German performance evolves as the training set grows from 2 to 100 examples (Italian results are qualitatively identical). The CNLMs already distinguish the categories well with small training sets, while the autoencoder does not catch up even with 100 training examples per category.

\begin{table}[t]
%  \begin{small}
\footnotesize
    \begin{center}
      \begin{tabular}{l|l|l}
        &\emph{German}&\emph{Italian}\\
        \hline
	      Random & 50.0 & 50.0 \\
        Autoencoder & 65.1 ($\pm$ 0.22) & 82.8 ($\pm$ 0.26) \\
	      \hline
        LSTM & 89.0 ($\pm$ 0.14) & 95.0 ($\pm$ 0.10) \\
        RNN & 82.0 ($\pm$ 0.64) & 91.9 ($\pm$ 0.24) \\
	      WordNLM & 53.5 ($\pm$ 0.18)  & 62.5 ($\pm$ 0.26) \\ \hline
	      WordNLM$_{\textit{subs.}}$ & 97.4 ($\pm$ 0.05) & 96.0 ($\pm$ 0.06) \\
      \end{tabular}
    \end{center}
 % \end{small}
	\caption{\label{tab:pos-results} Accuracy of diagnostic classifier on predicting word class, with standard errors across 100 random train-test splits.~`\emph{subs.}'~marks in-vocabulary subset evaluation, not comparable to the other results.} % Random accuracy is 50\%.} % (20 training examples)
\end{table}

% \begin{figure}
% \includegraphics[width=0.48\textwidth]{figures/german_pos_nouns_verbs.pdf}
% 	\caption{Word class accuracy as a function of training examples (German). }\label{fig:pos-induction}
% 	% \textbf{Please rename LM LSTM, Baseline Autoencoder and Words WordNLM}
% \end{figure}

\paragraph{Number}
We turn next to number, a more granular morphological feature. We
study German, as it possesses a rich system of nominal classes forming
plural through different morphological processes. We train a diagnostic number
classifier on a subset of these classes, and test on the others, in order to probe the abstract number generalization capabilities of the tested models. If a
model generalizes correctly, it means that the CNLM is sensitive to number
as an abstract feature, independently of its surface expression.

We extracted plural nouns from the Wiktionary and the German UD
treebank \cite{mcdonald2013universal,brants2002tiger}.  We
selected % de2006generating,
nouns with plurals in -\emph{n}, -\emph{s}, or -\emph{e} to train the
classifier (e.g., \emph{Geschichte(n)} `story(-ies)', \emph{Radio(s)}
`radio(s)', \emph{Pferd(e)} `horse(s)', respectively). We tested on
plurals formed with -\emph{r} (e.g., \emph{Lieder} for singular
\emph{Lied} `song'), or through vowel change (\emph{Umlaut}, e.g.,
\emph{{\"A}pfel} from singular \emph{Apfel} `apple'). Certain nouns
form plurals through concurrent suffixing and Umlaut. We grouped
these together with nouns using the same suffix, reserving the Umlaut
group for nouns \emph{only} undergoing vowel change (e.g.,
\emph{Saft/S\"afte} `juice(s)' would be an instance of -\emph{e}
suffixation). The diagnostic classifier was trained on 15 singulars and plurals
randomly selected from each training class.  As plural suffixes make
words longer, we sampled singulars and
plurals %used rejection sampling \textbf{(cite something?)}
from a single distribution over lengths, to ensure that their lengths
were approximately matched. Moreover, since in uncontrolled samples
from our training classes a final \emph{-e-} vowel would constitute a
strong surface cue to plurality, we balanced the distribution of this
property across singulars and plurals in the samples. For the test
set, we selected all plurals in -\emph{r} (127) or Umlaut (38), with
their respective singulars. %\textbf{(how many?)}.
We also used all remaining plurals ending in -\emph{n} (1467),
-\emph{s} (98) and -\emph{e} (832) as in-domain test data.
%\textbf{Need information on the test set for the training classes.}
To control for the impact of training sample selection, we report
accuracies averaged over 200 random train-test splits and
standard errors over these splits. %   We extract word representations as
% above, and we compare the same models.
For WordNLM OOV, there were 45.0\% OOVs in the training classes,
49.1\% among the -\emph{r} forms, and 52.1\% for Umlaut.

Results are in Table \ref{tab:number-results-e}. The classifier based
on word embeddings is the most successful. It outperforms in most cases
the best CNLM even in the more cogent OOV-inclusive evaluation. This
confirms the common observation that word embeddings reliably encode
number \cite{Mikolov:etal:2013a}. Again, the LSTM-based CNLM is better
than the RNN, but both significantly outperform the autoencoder. The
latter is near-random on new class prediction, confirming that we
properly controlled for orthographic confounds.

We observe a considerable drop in the LSTM CNLM performance between
generalization to -\emph{r} and Umlaut. On the one hand, the fact that
performance is still clearly above chance (and autoencoder) in the
latter condition shows that the LSTM CNLM has a somewhat abstract
notion of number not tied to specific orthographic exponents. On the
other, the -\emph{r} vs.~Umlaut difference suggests that the
generalization is not completely abstract, as it works more reliably
when the target is a new suffixation pattern, albeit one that is
distinct from those seen in training, than when
it is a purely non-concatenative process.

\begin{table}[t]
	\footnotesize
  \begin{center}
    \begin{tabular}{@{\hspace{0.3em}}l@{\hspace{0.42em}}|@{\hspace{0.42em}}l@{\hspace{0.45em}}|@{\hspace{0.45em}}l@{\hspace{0.65em}}l@{\hspace{0.15em}}}
      &train classes&\multicolumn{2}{c}{test classes}\\
      &\emph{-n/-s/-e}&\multicolumn{1}{c}{\emph{-r}}&\multicolumn{1}{c}{\emph{Umlaut}}\\      \hline
	    Random & 50.0 & 50.0 & 50.0 \\
	    Autoencoder & 61.4 ($\pm$ 0.9)  & 50.7 ($\pm$ 0.8)  & 51.9 ($\pm$ 0.4)  \\            \hline
	    LSTM & 71.5 ($\pm$ 0.8)  & 78.8 ($\pm$ 0.6)  & 60.8 ($\pm$ 0.6)  \\
	    RNN & 65.4 ($\pm$ 0.9)  & 59.8 ($\pm$ 1.0)  & 56.7 ($\pm$ 0.7)  \\
	    WordNLM  & 77.3 ($\pm$ 0.7)  & 77.1 ($\pm$ 0.5)  & 74.2 ($\pm$ 0.6)  \\ \hline
	    WordNLM$_{\textit{subs.}}$ & 97.1 ($\pm$ 0.3)  & 90.7 ($\pm$ 0.1)  & 97.5 ($\pm$ 0.1)  \\
    \end{tabular}
  \end{center}
  \caption{\label{tab:number-results-e} German number classification
	accuracy , with standard errors computed from 200 random train-test % when controlling for -\emph{e}-
	splits.~`\emph{subs.}'~marks in-vocabulary subset evaluation, not comparable to the other results.} % Random accuracy is 50\%.}
\end{table}

\subsection{Capturing syntactic dependencies}
\label{sec:dependencies}

Words encapsulate linguistic information into units that are then put
into relation by syntactic rules. A long tradition in linguistics has
even claimed that syntax is blind to sub-word-level processes
\cite[e.g.,][]{Chomsky:1970,DiSciullo:Williams:1987,Bresnan:Mchombo:1995,Williams:2007}. Can
our CNLMs, despite the lack of an explicit word lexicon, capture
relational syntactic phenomena, such as agreement and case assignment?
We investigate this by testing them on syntactic dependencies between
non-adjacent words. We adopt the ``grammaticality judgment'' paradigm
of \newcite{Linzen:etal:2016}. We create minimal sets of grammatical
and ungrammatical phrases illustrating the phenomenon of interest, and
let the language model assign a likelihood to all items in the
set. The language model is said to ``prefer'' the grammatical variant
if it assigns a higher likelihood to it than to its ungrammatical
counterparts. We must stress two methodological points. First, since a
character-level language model assigns a probability to each character
of a phrase, and the phrase likelihood is the product of these values
(all between 0 and 1), minimal sets must be controlled for character
length. This makes existing benchmarks unusable. Second, the
``distance'' of a relation is defined differently for a
character-level model, and it is not straightforward to
quantify. Consider the German phrase in \ref{ex:german-gender}
below. For a word model, two items separate the article from the
noun. For a (space-less) character model, 8 characters intervene until
the noun onset, but the span to consider will typically be longer. For
example, \emph{Baum} could be the beginning of the feminine noun
\emph{Baumwolle} `cotton', which would change the agreement
requirements on the article. So, until the model finds evidence that
it fully parsed the head noun, it cannot reliably check
agreement. This will typically require parsing at least the full noun
and the first character following it. We again focus on German and Italian, as their richer inflectional
morphology simplifies the task of constructing balanced minimal sets.

%\textbf{Please provide size for all evaluation sets.}

%\textbf{In the interest of space, please reduce figures ~\ref{fig:gender}, ~\ref{fig:case} and \ref{fig:prep} to a single figure with 3 panels: the averaged gender and case results, and the subcategorization case. Also, either put titles on the figures, or at least label them as a), b) c). Legends should be LSTM, RNN and WordNLM (or WNLM) for coherence.}XS

%We take a further step up the linguistic hierarchy, probing CNLMs for their ability to capture syntactic dependencies between non-adjacent words. %
% --a rather challenging task for models that work entirely at the character level, and do not even have information about what words are.
% % Despite not having predefined information about words and
% % morphemes, is the model able to capture non-adjacent syntactic
% % dependencies?
% In particular, is it able to do so when dependencies cross one or more words, and thus cannot be reduced to surface n-gram counts?
% Note that, for a CNLM, dependencies across even a single word are often already long-distance. % even \emph{``\textbf{la} bell\textbf{a}''} is long distance.
% We again focus on German and Italian due to the richness of inflectional morphology in these languages.
% Constructions will be language-specific, so we discuss the languages separately. %German and Italian separately (not much in English).

%As usual, specifics of training etc that depart from general setup.

\subsubsection{German}%  We consider 4 constructions:
% \begin{inparaenum}[i)]
% \item article-noun gender agreement, possibly with material in the middle,
% \item determiner-noun case concord, again with material in the middle,
% \item preposition case sub-categorization, with material in the middle.
% \end{inparaenum}

\paragraph{Article-noun  gender agreement}
Each German noun belongs to one of three genders (masculine, feminine, neuter), morphologically marked on the article. As the article and the noun can be separated by adjectives and adverbs, we can probe knowledge of lexical gender together with long-distance agreement.
We create stimuli of the form
\exg. \{\underline{der},\ die,\ das\} sehr rote Baum \\
the very red tree \label{ex:german-gender}\\
%    article adverb adjective noun\\
%    `the very red tree'

%\begin{enumerate}[label={(\arabic*)}]
%	\item \begin{tabular}[t]{lllllll}
%	\{\underline{der}, die, das\}& sehr& rote& Baum \\
%	article & adverb & adjective & noun \\
%	the & very & red & tree
%\end{tabular}
%\end{enumerate}
where the correct nominative singular article (\emph{der}, in this
case) matches the gender of the noun.  We then run the CNLM on the
three versions of this phrase (removing whitespace) and record the
probabilities it assigns to them. If the model assigns the highest
probability to the version with the right article, we count it as a
hit for the model. To avoid phrase segmentation ambiguities (as in the
\emph{Baum}/\emph{Baumwolle} example above), we present phrases
surrounded by full stops.

%  \cite{de2006generating,mcdonald2013universal}
To build the test set, we select all 4,581 nominative singular nouns
from the German UD treebank: 49.3\% feminine, 26.4\% masculine, 24.3\% neuter. WordNLM OOV noun ratios are: 40.0\% for masculine, 36.2\%
for feminine, 41.5\% for neuter.  We construct four conditions
varying the number of adverbs and adjectives between article and noun.
We first consider stimuli where no material
intervenes. % \footnote{Due to syncretism in the article paradigm, there is sometimes ambiguity in the choice of the correct article if the noun's morphology does not uniquely indicate that it is a nominative singular from. As this equally affects all feminine nouns, we did not remove such cases. Importantly, this issue is solved as soon as an adjective intervenes, as its form disambiguates case.}
In the second condition, an adjective with the correct case ending,
randomly selected from the training corpus, is added. Crucially, the
ending of the adjective does not reveal the gender of the noun.  We
only used adjectives occurring at least 100 times, and not ending in
-\emph{r}.\footnote{Adjectives ending in -\emph{r} often reflect lemmatization provblems, as TreeTagger occasionally failed to remove the
  inflectional suffix -\emph{r} when lemmatizing. We needed to extract lemmas, as we constructed the appropriate inflected forms on their basis.}  We obtained a pool
of 9,742 adjectives to sample from, also used in subsequent
experiments.  74.9\% of these were OOV for the WordNLM.  In the third
and fourth conditions, one (\emph{sehr}) or two adverbs (\emph{sehr
  extrem}) intervene between article and adjective. These do not cue
gender either. We obtained 2,290 (m.), 2,261 (f.), and 1,111 (n.)
stimuli, respectively. To control for surface co-occurrence statistics
in the input, we constructed an n-gram baseline picking the article
most frequently occurring before the phrase in the training data,
breaking ties randomly. OOVs were excluded from WordNLM evaluation,
resulting in an easier test for this rival model. However, here and in
the next two tasks, CNLM performance on this reduced set was only
slightly better, and we do not report it here.
We report accuracy averaged over nouns belonging to each of the three genders.
By design, the random baseline accuracy is 33\%. % Marco: I added this, can you check it's clear?

\begin{figure*}
	\begin{tabular}{cccc}
		& Gender & Case & Subcategorization \\ 
		\raisebox{1.7\height}{\rotatebox[origin=c]{90}{Accuracy}}
		&
\includegraphics[width=0.28\textwidth]{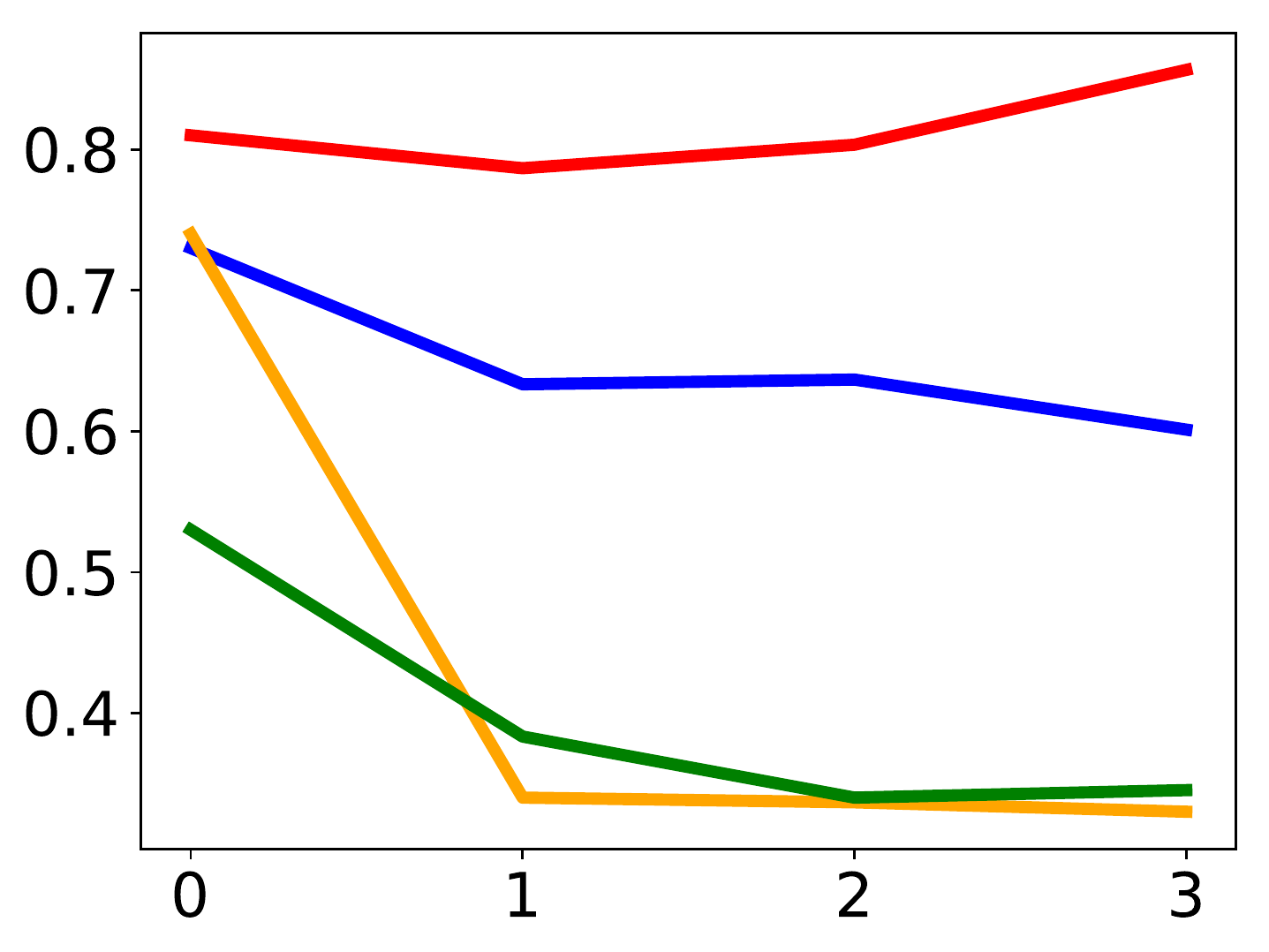} 
		&
		\includegraphics[width=0.28\textwidth]{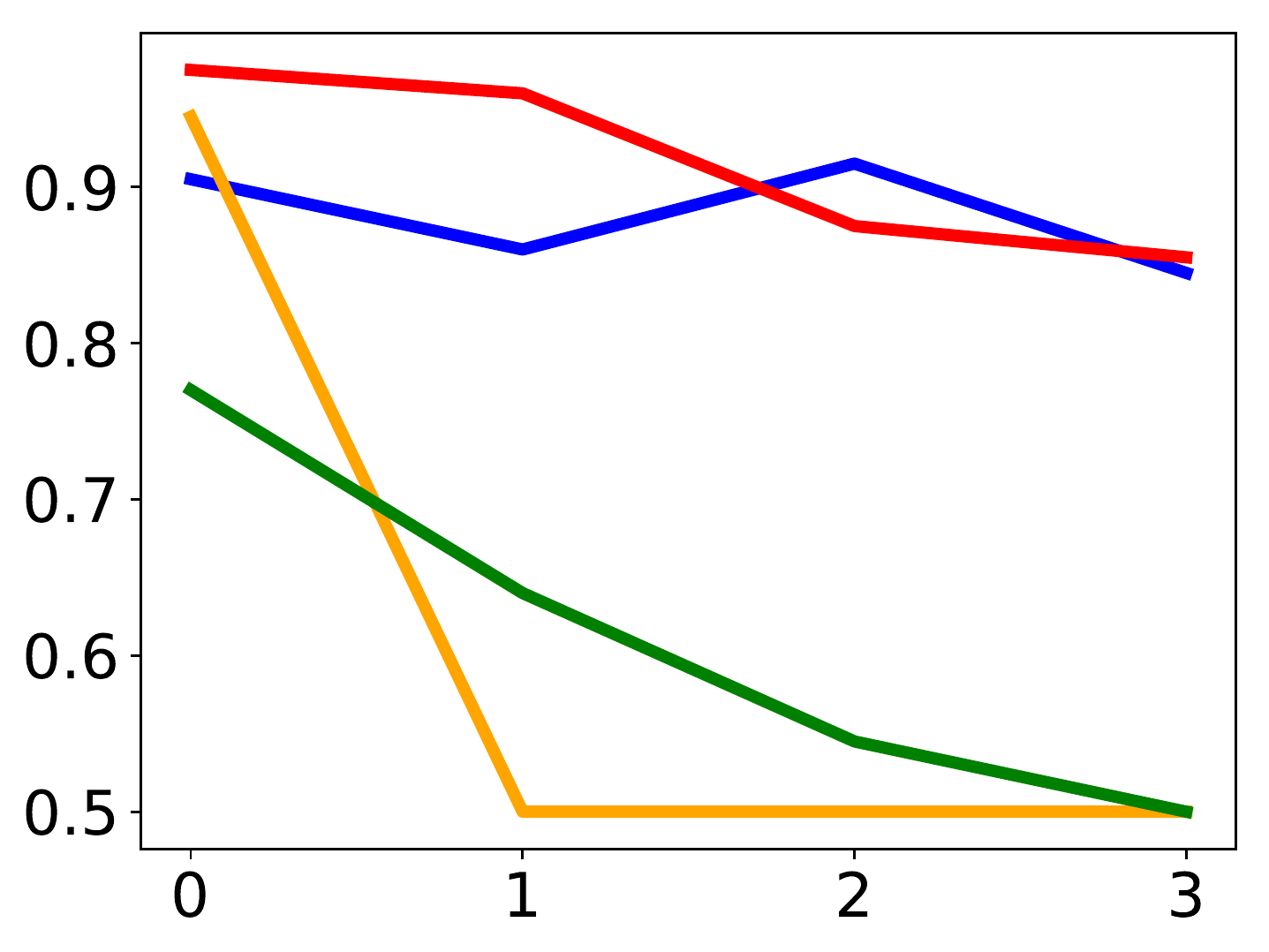}
		&
\includegraphics[width=0.28\textwidth]{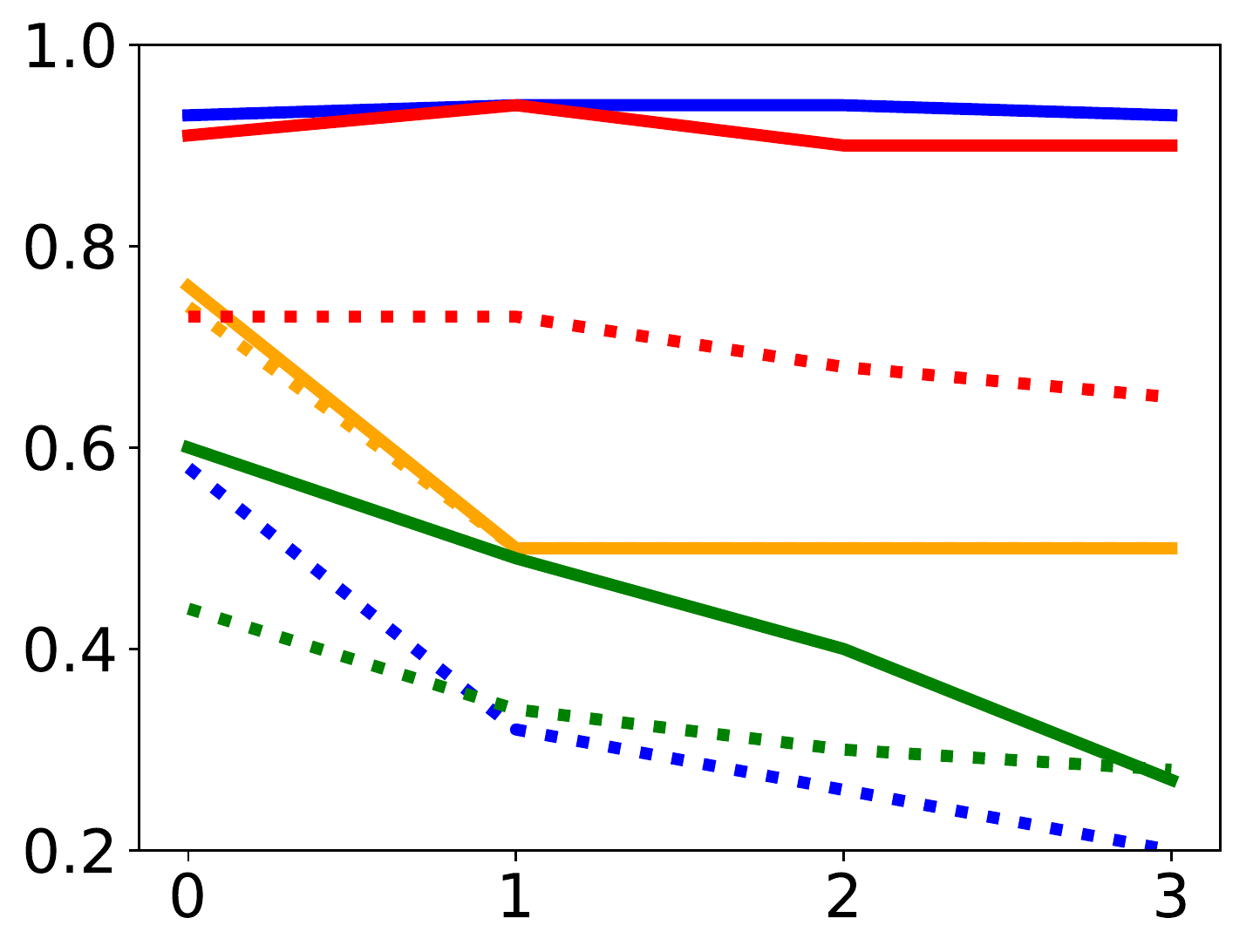} \\ 
		&\multicolumn{3}{c}{Number of Intervening Words}
	\end{tabular}
\centering\includegraphics[width=0.5\textwidth]{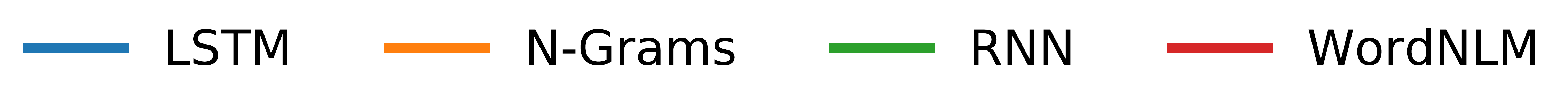}
\caption{Accuracy in the German syntax tasks, in function of number of intervening words.}\label{fig:german-syntax}
\end{figure*}

Results are presented in Figure \ref{fig:german-syntax}
(left). WordNLM performs best, followed by the LSTM CNLM.  The n-gram
baseline performs similarly to the CNLM when there is no intervening
material, which is expected, as a noun will often be preceded by its
article in the corpus. However, its accuracy drops to chance level
(0.33) in the presence of an adjective, whereas the CNLM is still able
to track agreement. %
% This problem would not be
% mitigated by interpolation with or backoff to lower-order n-grams, as
% the relevant gender information is present only on the first and last
% word of each stimulus. We conclude that, while direct association
% between articles and nouns can be learnt from simple corpus
% statistics, the CNLM has some capability to preserve the relevant
% information across more than a dozen timesteps.
The RNN variant is much worse. It is outperformed by the n-gram model
in the adjacent condition, and it drops to random accuracy as more
material intervenes. We emphasized at the outset of this section that
CNLMs must track agreement across much wider spans than
word-based models. The LSTM variant ability to preserve information for
longer might play a crucial role here.

% Note that, at the character level, even ``adjacent''
% agreement requires carrying information through multiple time steps
% (the agreement violation will not emerge until enough characters of
% the noun have been processed to disambiguate its gender with respect
% to its prefix-sharing cohort).

% The exclusion of OOVs and thus limiting experiments on word-level models to frequent words might create an unfair advantage; running the CNLM only on those stimuli given to the word-level model results in slightly better accuracies but the same pattern of results.

% \begin{figure*}
% % \includegraphics[width=0.24\textwidth]{figures/german-gender-m.pdf}
% % \includegraphics[width=0.24\textwidth]{figures/german-gender-f.pdf}
% % \includegraphics[width=0.24\textwidth]{figures/german-gender-n.pdf}
% 	\begin{tabular}{ccc}
% Gender & Case & Subcategorization \\
% \includegraphics[width=0.33\textwidth]{figures/german-gender-total.pdf} 
% 		&
% 		\includegraphics[width=0.33\textwidth]{figures/german-case-total.pdf}
% 		&
% \includegraphics[width=0.33\textwidth]{figures/german-prep-with-control.pdf}
% 	\end{tabular}
% \centering\includegraphics[width=0.5\textwidth]{figures/german-legend.pdf}
% \caption{Accuracy on the German syntax tasks, as a function of the number of intervening elements.}\label{fig:german-syntax}
% \end{figure*}

\paragraph{Article-noun case agreement}
We selected the two determiners \emph{dem} and \emph{des}, which
unambiguously indicate dative and genitive case, respectively, for
masculine and neuter nouns: %\textbf{Give one example.}
\ex.\ag. {\{\underline{dem}, des\}} sehr roten Baum \\
the very red {tree \emph{(dative)}} \\
\bg. {\{dem, \underline{des}\}} sehr roten Baums \\
the very red {tree \emph{(genitive)}} \\

%\begin{enumerate}[label={(\arabic*)}]
%	\item \begin{tabular}[t]{lllllll}
%	\{\underline{dem}, des\}& sehr& roten& Baum \\
%	article & adverb & adjective & noun \\
%	the & very & red & tree
%\end{tabular}
%\item \begin{tabular}[t]{lllllll}
%	\{dem, \underline{des}\}& sehr& roten& Baums \\
%	article & adverb & adjective & noun \\
%	the & very & red & tree
%\end{tabular}
%
%\end{enumerate}

We selected all noun lemmas of the appropriate genders from the German
UD treebank, and extracted morphological paradigms from Wiktionary to
obtain case-marked forms, retaining only nouns unambiguously marking
the two cases (4,509 nouns).  We created four conditions, varying the amount of
intervening material, as in the gender agreement experiment (4,509
stimuli per condition).  For 81.3\% of the nouns, at least one of the
two forms was OOV for the WordNLM, and we tested the latter on the
full-coverage subset.
Random baseline accuracy is 50\%.

Results are in Figure \ref{fig:german-syntax} (center).  Again,
WordNLM has the best performance, but the LSTM CNLM is competitive as
more elements intervene. Accuracy stays well above 80\% even with
three intervening words.  The n-gram model performs well if there is
no intervening material (again reflecting the obvious fact that
article-noun sequences are frequent in the corpus), and at chance
otherwise.  The RNN CNLM accuracy is above chance with one and two
intervening elements, but drops considerably with distance.

% Considering the results for the dative and genitive separately, accuracy slightly increases in the dative case and decreases in the genitive case.
% This can be attributed to the higher baseline frequency of dative in German, suggesting that both word- and character-based networks are impacted by unigram frequencies as more words intervene.
% This effect is far more pronounced for the RNN CNLM, explaining its overall decrease to chance level.
% Again, restricting to words that are in the word-level vocabulary did not change the pattern of results.
%\begin{figure}
%% \includegraphics[width=0.23\textwidth]{figures/german-case-Dative.pdf}
%% \includegraphics[width=0.23\textwidth]{figures/german-case-Genitive.pdf} \\
%
%\centering\includegraphics[width=0.24\textwidth]{figures/german-case-total.pdf}
%
%\includegraphics[width=0.5\textwidth]{figures/german-legend.pdf}
%	\caption{\textbf{(Reduce to single figure, 2)} Accuracy on the case agreement task as a function of the number of intervening elements.}\label{fig:case}
%\end{figure}

\paragraph{Prepositional case subcategorization}
German verbs and prepositions lexically specify their object's case.  We
study the preposition \textit{mit} `with', which selects a dative
object. We focus on \textit{mit}, as it unambiguously requires a dative object,
and it is extremely frequent in the Wikipedia corpus we are using. To build the test set,
we select objects whose head noun is a nominalized adjective,
with regular, overtly marked case inflection.
We use the same adjective pool as in the preceding experiments.
%We take all adjectives
%that occur at least 100 times in the training data, excluding those
%that end in -\emph{r}, as these often reflected lemmatization
%problems.
We then select all sentences containing a \emph{mit} prepositional
phrase in the German Universal Dependencies treebank, subject to the
constraints that (1) the head of the noun phrase governed by the
preposition is not a pronoun (replacing such items with a nominal
object often results in ungrammaticality), and (2) the governed noun
phrase is continuous, in the sense that it is not interrupted by words
that do not belong to it.\footnote{The main source of noun phrase
  discontinuity in the German UD corpus is extraposition, a common
  phenomenon where part of the noun phrase is separated from the rest
  by the verb.} %\textbf{(please explain the latter more clearly)}.
We obtained 1,629 such sentences.  For each sentence, we remove the
prepositional phrase and replace it by a phrase of the form
\exg. mit der sehr \{rote,\ \underline{roten}\} \\
with the very red\ one \\

%\begin{enumerate}[label={(\arabic*)}]
%	\item \begin{tabular}[t]{lllllll}
%	mit & der & sehr& \{rote, \underline{roten}\} \\
%	prep & article  & adverb & adjective \\
%	with & the & very  & red one 
%\end{tabular}
%\end{enumerate}
where only the \emph{-en} (dative) version of the adjective is
compatible with the case requirement of the preposition (and the
intervening material does not disambiguate case). We construct three
conditions by varying the presence and number of adverbs (\emph{sehr}
`very', \emph{sehr extrem} `very extremely', \emph{sehr extrem
  unglaublich} `very extremely incredibly').  Note that here the
correct form is longer than the wrong one. As the overall likelihood
is the product of character probabilities ranging between 0 and 1, if
this introduces a length bias, the latter will work against the
character models. % ~\cite{sountsov2016length}.
Note also that we embed test phrases into full sentences (e.g., \emph{Die Figur hat \textbf{mit der roten} gespielt und meistens gewonnen.} `The figure played with the red one and mostly won').
We
do this %as sentence-initial \emph{mit} is somewhat unnatural and, more importantly,
because this will disambiguate the final element of the
phrase as a noun (not an adjective), and exclude the reading in which
\emph{mit} is a particle not governing the noun phrase of
interest~\citep{duden2019mit}.\footnote{An example of this unintended reading of \emph{mit} is: \textit{Ich war \textbf{mit} der erste, der hier war.} `I was one of the first who arrived here.' In this context, dative \emph{ersten} would be ungrammatical.}
When running the WordNLM, we excluded OOV adjectives as in
the previous experiments, but did not apply further OOV filtering to
the sentence frames.  For the n-gram baseline, we
only counted occurrences of the prepositional phrase, omitting the
sentential contexts.
Random baseline accuracy is 50\%.
% A full stimulus example with the wrapping sentence is:
% \exg. Die Figure hat mit der \{rote,\ \underline{roten}\} gespielt und meistens gewonnen . \\
% the figure has with the red\ one played and mostly won \\
% \textit{`The figure played with the red one and mostly won.'}

%\textbf{Add a full stimulus example (with the wrapping
%  sentence).} \textbf{OOV statistics.}

% Note also that in this case we embedded the test phrase into full sentences.
% We did this as sentences beginning with \emph{mit} are somewhat unnatural
% and, more importantly since \emph{mit} is homophonous with a particle
% that does not select for an object, and so that the following context disambiguates
% the last word in the phrase as a nominalization (as opposed to an adjective).

We also created control stimuli where all words up to and including
the preposition are removed (the example sentence above becomes:
\textit{der roten gespielt und meistens gewonnen}). If a model's
accuracy is lower on these control stimuli than on the full ones, its
performance cannot be simply explained by the different unigram
probabilities of the two adjective forms.
%\begin{figure}
%\includegraphics[width=0.48\textwidth]{figures/german-prep-with-control.pdf}
%
%\includegraphics[width=0.48\textwidth]{figures/german-legend.pdf}
%\caption{\textbf{(Reduce to single figure, 3)} Accuracy on the subcategorization task as a function of the number of intervening elements. The dashed lines indicate accuracies on the control stimuli that do not contain the preposition.}\label{fig:prep}
%\end{figure}
%

Results are shown in Figure \ref{fig:german-syntax} (right). Only the
n-gram baseline fails to outperform control accuracy (dotted). Surprisingly,
the LSTM CNLM slightly outperforms the WordNLM, even though the latter
is evaluated on the easier full-lexical-coverage stimulus subset.
Neither model shows accuracy decay as the number of adverbs increases.
As before, the n-gram model drops to chance as adverbs intervene,
while the RNN CNLM starts with low accuracy that progressively decays
below chance.

%Results are shown in Table~\ref{tab:ital-agr-results}.
%The word LSTM shows the highest overall performance, closely followed by the LSTM CNLM.
%The RNN performs well on adjective gender, and considerably worse than the CNLM on the other tasks.
%For the CNLMs, the most challenging task was article-noun gender agreement.
%Discussion case-by-case, including how we control for n-gram frequency
%and length.
%Results table with a row for each pattern and a column for each model.

\subsubsection{Italian} %\textbf{Also here, add data-set sizes.}
%\textbf{For the examples, you know the linguex packet, right?}
%\textbf{Given that you itemize by gender, can you add a baseline based
%  on picking the most frequent variant of the noun/adjective?}
% We focus on paradigms where gender and number are explicitly and
% systematically encoded and it is possible to compare same-length
% strings. We are able to extract enough stimuli that never occur in the
% training corpus, so that an n-gram control would be at chance
% level. Moreover, by experiment construction, baselines relying on
% unigram frequency are also at chance level.

\paragraph{Article-noun gender agreement}
%(1) eadj-aonoun:

Similar to German, Italian articles agree with the noun in gender;
however, Italian has a relatively extended paradigm of masculine and
feminine nouns differing only in the final vowel (-\emph{o} and
-\emph{a}, respectively), allowing us to test agreement in fully
controlled paradigms such as the following:

\ex.\ag. \{\underline{il},\ la\}  congeniale  candidato \\
the congenial candidate\ (m.) \\
\bg.  \{il,\ \underline{la}\}  congeniale  candidata \\
the congenial candidate\ (f.) \\
%\begin{enumerate}[label={(\arabic*)}]
%	\item 
%		\begin{tabular}[t]{lllllll}
%	a. & \{\underline{il}, la\} & congeniale & candidato \\
%   &  the & congenial & candidate (m.) \\
%	& \multicolumn{4}{l}{`The congenial male candidate.'} \\
%	b. & \{il, \underline{la}\} & congeniale & candidata \\
%    &the & congenial & candidate (f.) \\
%	& \multicolumn{4}{l}{`The congenial female candidate.'} \\
%\end{tabular}
%\end{enumerate}

The intervening adjective, ending in -\emph{e}, does not cue
gender. We constructed the stimuli with words appearing at least 100
times in the training corpus. We required moreover the \emph{-a}
and \emph{-o} forms of a noun to be reasonably balanced in frequency
(neither form is more than twice as frequent as the other), or both rather
frequent (appear at least 500 times). As the prenominal adjectives are
somewhat marked, we only considered -\emph{e} adjectives that occur
prenominally with at least 10 distinct nouns in the training corpus.
Here and below, stimuli were manually checked removing nonsensical
adjective-noun (below, adverb-adjective) combinations. Finally, adjective-noun combinations that
occurred in the training corpus were excluded, so that an n-gram
baseline would perform at chance level. We obtained 15,005 stimulus
pairs in total. % In
% addition to CNLMs and WordNLM, we tested frequency baseline choosing
% the option with the higher unigram probability (whole-stimulus corpus
% frequencies are at chance by construction).
35.8\% of them contained an adjective or noun that was
OOV for the WordNLM. Again, we report this model's results on its
full-coverage subset, where the CNLM performance is only slightly above
the one reported.

Results are shown on the first line of
Table \ref{tab:ital-agr-results}.  WordNLM shows the strongest
performance, closely followed by the LSTM CNLM.  The RNN CNLM
performs strongly above chance (50\%), but again lags behind the LSTM.

\paragraph{Article-adjective gender agreement}
We next consider agreement between articles and adjectives with an intervening adverb:
\ex.\ag. il meno \{\underline{alieno},\ aliena\} \\
the\ (m.) less alien\ one \\
\bg. la meno \{alieno,\ \underline{aliena}\} \\
the\ (f.) less alien\ one \\

%\begin{enumerate}[label={(\arabic*)}]
%	\item 
%		\begin{tabular}[t]{lllllll}
%	a. & il & meno & \{ \underline{alieno}, aliena \} \\
%   &  the (m.)& less & alien  \\
%	b. & la & meno & \{ alieno, \underline{aliena} \} \\
%    &the (f.)& less & alien one \\
%\end{tabular}
%\end{enumerate}
where we used the adverbs \emph{pi{\`u}} `more', \emph{meno} `less',
\emph{tanto} `so much'. We considered only adjectives that occurred 1K
times in the training corpus (as adjectives ending in \emph{-a}/\emph{-o} are
very common). We excluded all cases in which the
adverb-adjective combination occurred in the training corpus, obtaining 88 stimulus pairs. Due to the restriction to common adjectives, there were no WordNLM OOVs. %
%
%Here and in the next experiment, the frequency baseline chose the version with the more common adjective.
% /checkpoint/mbaroni/char-rnn-exchange/candidate_adv_aoadj_testset.txt
Results are shown on the second line of
Table~\ref{tab:ital-agr-results}; all three models perform almost
perfectly.  Possibly, the task is made easy by the use of extremely common
adverbs and adjectives.

\begin{table}[t]
  \begin{small}
    \begin{center}
      \begin{tabular}{l|ll|c}
        & \multicolumn{2}{c|}{CNLM} & \multicolumn{1}{c}{\multirow{2}{*}{WordNLM}}\\
        &\emph{LSTM}&\emph{RNN} &  \\ \hline
% eadj-aonoun
        Noun Gender & 93.1  & 79.2 & 97.4\\
%      adv-aoadj
        Adj.~Gender & 99.5 & 98.9 & 99.5\\
% adv-aeadj
        Adj.~Number & 99.0 & 84.5 & 100.0\\
%	    & \multicolumn{4}{c|}{CNLM} & \multicolumn{2}{c|}{\multirow{2}{*}{WordNLM}}  & \multicolumn{2}{c}{\multirow{2}{*}{Frequency}}\\
%	    &\multicolumn{2}{c|}{\emph{LSTM}}&\multicolumn{2}{c|}{\emph{RNN}} & & & \\ \hline
%% eadj-aonoun
%	    Noun Gender & 96.6&89.6  & 84.4&73.9 & 99.1&95.6 & 100.0 & 0.0\\
%%      adv-aoadj
%	    Adj.~Gender & 98.9&100.0 & 100.0&97.8 & 98.9&100.0 & 55.7 & 44.3 \\
%% adv-aeadj
%	    Adj.~Number & 99.0&99.0 & 99.0&70.0 & 100.0&100.0 & 86.7 & 13.3 \\
%
      \end{tabular}
    \end{center}
  \end{small}
  \caption{\label{tab:ital-agr-results} Italian agreement results. Random baseline accuracy is 50\% in all three experiments.} % For each model and test, we report percentage accuracy on two stimulus classes (masculine/feminine for gender, singular/plural for number).}
\end{table}

% \begin{table}[t]
%   \begin{small}
%     \begin{center}
%       \begin{tabular}{l|ll|l|l}
%         & \multicolumn{2}{c|}{CNLM} & \multicolumn{1}{c|}{\multirow{2}{*}{WordNLM}}  & \multicolumn{1}{c}{\multirow{2}{*}{Freq.}}\\
%         &\emph{LSTM}&\emph{RNN} &  \\ \hline
% % eadj-aonoun
%         Noun Gender & 93.1  & 79.2 & 97.4 & 50.0\\
% %      adv-aoadj
%         Adj.~Gender & 99.5 & 98.9 & 99.5 & 50.0\\
% % adv-aeadj
%         Adj.~Number & 99.0 & 84.5 & 100.0 & 50.0 \\
% %	    & \multicolumn{4}{c|}{CNLM} & \multicolumn{2}{c|}{\multirow{2}{*}{WordNLM}}  & \multicolumn{2}{c}{\multirow{2}{*}{Frequency}}\\
% %	    &\multicolumn{2}{c|}{\emph{LSTM}}&\multicolumn{2}{c|}{\emph{RNN}} & & & \\ \hline
% %% eadj-aonoun
% %	    Noun Gender & 96.6&89.6  & 84.4&73.9 & 99.1&95.6 & 100.0 & 0.0\\
% %%      adv-aoadj
% %	    Adj.~Gender & 98.9&100.0 & 100.0&97.8 & 98.9&100.0 & 55.7 & 44.3 \\
% %% adv-aeadj
% %	    Adj.~Number & 99.0&99.0 & 99.0&70.0 & 100.0&100.0 & 86.7 & 13.3 \\
% %
%       \end{tabular}
%     \end{center}
%   \end{small}
%   \caption{\label{tab:ital-agr-results} Italian agreement results.} % For each model and test, we report percentage accuracy on two stimulus classes (masculine/feminine for gender, singular/plural for number).}
% \end{table}

\paragraph{Article-adjective number agreement}
Finally, we constructed a version of the last test that probed number agreement. For feminine forms, it is possible to compare same-length phrases such as:
\ex.\ag. la meno \{\underline{aliena},\ aliene\} \\
the\ (s.) less alien\ one(s) \\
\bg. le meno \{aliena,\ \underline{aliene}\} \\
the\ (p.) less alien\ one(s) \\

%\begin{enumerate}[label={(\arabic*)}]
%	\item 
%\begin{tabular}[t]{lllllll}
%	a. & la & meno & \{ \underline{aliena}, aliene \} \\
%   &  the (s.)& less & alien one(s)  \\
%	b. & le & meno & \{ aliena, \underline{aliene} \} \\
%    &the (p.)& less & alien one(s) \\
%\end{tabular}
%\end{enumerate}
% /checkpoint/mbaroni/char-rnn-exchange/candidate_adv_aeadj_testset.txt
Stimulus selection was as in the last experiment, but we used a 500-occurrences
threshold for adjectives, as feminine plurals are less common, obtaining 99 pairs. Again, no adverb-adjective combination was attested. %  Further, we manually
There were no OOV items for the WordNLM. %
% removed adjectives that did not combine well semantically with the
% adverbs under consideration (\emph{pi{\`u}, meno, tanto}).
Results are shown on the third line of
Table~\ref{tab:ital-agr-results}; the LSTMs perform almost perfectly,
and the RNN is strongly above chance.

\subsection{Semantics-driven sentence completion}
\label{sec:semantics}

We probe whether CNLMs are capable of tracking the shallow
form of word-level semantics required in a fill-the-gap test. We turn now to English, as for this language we can use the Microsoft
Research Sentence Completion task \cite{Zweig:Burges:2011}. The
challenge consists of sentences with a gap, with 5 possible
choices to fill it. Language models can be
directly applied to the task, by calculating the likelihood of sentence
variants with all possible completions, and selecting the one with the
highest likelihood.

The creators of the benchmark took multiple precautions to insure that
success on the task implies some command of semantics. The multiple
choices were controlled for frequency, and the annotators were
encouraged to choose confounders whose elimination required ``semantic
knowledge and logical inference'' \cite{Zweig:Burges:2011}.  For
example, the right choice in \emph{``Was she his [
  \underline{client}|musings|discomfiture|choice|opportunity], his
  friend, or his mistress?} depends on the cue that the missing word is
coordinated with \emph{friend} and \emph{mistress}, and the
latter are animate entities.

The task domain (Sherlock Holmes novels) is very different
from the Wikipedia data-set we originally trained our models on. For a
fairer comparison with previous work, we re-trained our models on
the corpus provided with the benchmark, consisting of
41 Million words from 19th century English novels (we removed
whitespace from this corpus as well).
% thus we additionally trained our models on the training set provided
% for the task, consisting of 19th century English novels.
% %We both consider a fresh model trained on
%that data, and initializing it with the Wikipedia model.
% \footnote{For the WordNLM, the vocabulary consisted
% of the 50,000 most common words in the in-domain training set.}
%For comparison, we report results (KN5 from , LSTM from ) from previous work that were trained on the 19th century novels dataset (but the LSTM from that work had Glove embeddings). % \cite{zhang2016top} has a nice table if we want to report more

Results are in Table~\ref{tab:msr-completion-results}.  We
confirm the importance of in-domain training, as the models
trained on Wikipedia perform poorly (but still above chance level,
which is at 20\%).  With in-domain training, the LSTM CNLM outperforms
many earlier word-level neural models, and is only slightly below our
WordNLM. %, and approaches the best published results.
%, held by approaches developed for the completion task \cite{woods2016exploiting}.
% The best results I could find, https://github.com/ctr4si/sentence-completion, are much better than the best peer-reviewed published ones
The RNN is not successful even when trained in-domain,
contrasting with the \emph{word}-based vanilla RNN from the
literature, whose performance, while still below LSTMs, is much
stronger. Once more, this suggests that capturing word-level generalizations with a word-lexicon-less character model requires the long-span processing abilities of an LSTM.

\begin{table}[t]
  \footnotesize{
    \begin{center}
      \begin{tabular}{l|l}
        % \multicolumn{1}{c}{}& Model \\
        \multicolumn{2}{c}{\emph{Our models (wiki/in-domain)}}\\
\hline
        LSTM 	    &      34.1/59.0 \\ % /59.2
        RNN  &     24.3/24.0 \\ % /27.1
        WordNLM & 37.1/63.3 \\ %\hline % 50.1/52.4/
%        Random & 20.0 \\ \hline
\end{tabular}

\begin{tabular}{l|l||l|l}
%\hline
  \multicolumn{4}{c}{\emph{From the literature}}\\
  \hline
  KN5   & 40.0            & Skipgram         & 48.0    \\                                
        Word RNN & 45.0         & Skipgram + RNNs  & 58.9 \\                                  
        Word LSTM & 56.0        & PMI &  61.4 \\                      
        LdTreeLSTM  & 60.7     &  Context-Embed & 65.1 \\       \hline
             \end{tabular}
    \end{center}
  }
	\caption{\label{tab:msr-completion-results} Results on MSR Sentence Completion. For our models (top), we show accuracies for  Wikipedia (left) and in-domain (right) training. We compare with language models from prior work (left): Kneser-Ney 5-gram model \cite{Mikolov:2012}, Word RNN \cite{zweig2012computational}, Word LSTM and LdTreeLSTM \cite{zhang2016top}. We further report models incorporating distributional encodings of semantics (right): Skipgram(+RNNs) from \newcite{DBLP:journals/corr/abs-1301-3781}, the PMI-based model of \citet{woods2016exploiting}, and the Context-Embedding-based approach of \citet{melamud2016context2vec}.}
\end{table}

%
%
%\begin{table*}[t]
%  \begin{small}
%    \begin{center}
%      \begin{tabular}{l|llllllllllllllllll}
%        % \multicolumn{1}{c}{}& Model \\
%        LSTM 	    &      34.1/59.0  &     KN5   & 40.0          & Skipgram + RNNs  & 58.9 \\           
%        RNN  &     24.3/24.0          &     Word RNN & 45.0       & \citet{woods2016exploiting} &  61.4 \\ 
%        WordNLM & 37.1/63.3           &     Word LSTM & 56.0      & \citet{melamud2016context2vec} & 65.1 \\ 
%                &                     &      LdTreeLSTM  & 60.67  &                 \\
%             \end{tabular}
%    \end{center}
%  \end{small}
%  \caption{\label{tab:msr-completion-results} Results on MSR Sentence Completion. For our models, we show accuracies for  Wikipedia/in-domain training. We compare with language models from prior work: Kneser-Ney 5-gram model \cite{Mikolov:2012}, Word RNN \cite{zweig2012computational}, Word LSTM and LdTreeLSTM \cite{zhang2016top}. We further report models incorporating distributional encodings of semantics: Skipgram+RNNs from \newcite{Mikolov:etal:2013b}, the PMI-based model of \citet{woods2016exploiting}, and the context-embedding based approach by \citet{melamud2016context2vec}.}
%\end{table*}

%, (3) the Wikipedia model posttrained on the in-domain data
% For the WordNLM, we additionally provide accuracy for a model with vocabulary derived from the in-domain training data. 
% \textbf{Explain what these are}

\subsection{Boundary tracking in CNLMs}
\label{sec:segmentation}

The good performance of CNLMs on most tasks above suggests that,
although they lack a hard-coded word vocabulary and they were trained
on unsegmented input, there is enough pressure from the language
modeling task for them to learn to track word-like items, and
associate them with various morphological, syntactic and semantic
properties. In this section, we take a direct look at \emph{how} CNLMs
might be segmenting their input.
\newcite{Kementchedjhieva:Lopez:2018} found a \emph{single} unit in
their English CNLM that seems, qualitatively, to be tracking
morpheme/word boundaries.  Since they trained the model with
whitespace, the main function of this unit could simply be to predict
the very frequent whitespace character. We conjecture instead (like them) that the
ability to segment the input into meaningful items is so important
when processing language that CNLMs will specialize units for boundary
tracking even when trained without whitespace.

To look for ``boundary units'', we created a random set of 10,000
positions from the training set, balanced between those corresponding
to a word-final character and those occurring word-initially or
word-medially.  We then computed, for each hidden unit, the Pearson
correlation between its activations and a binary variable that takes value 1 in
word-final position and 0 elsewhere. %dummy variable coding
%word-final position.
For each language and model (LSTM or RNN), we found very few units
with a high correlation score, suggesting that the models have indeed
specialized units for boundary tracking. We further study the units
with the highest correlations, that are, for the LSTMs, 0.58
(English), 0.69 (German), and 0.57 (Italian).  For the RNNs, the
highest correlations are 0.40 (English), 0.46 (German and
Italian).\footnote{In an early version of this analysis, we
  arbitrarily imposed a minimum 0.70 correlation threshold, missing
  the presence of these units. We thank the reviewer who encouraged us to
  look further into the matter.}

\begin{figure*}
  \begin{center}
  \includegraphics[width=0.9\textwidth]{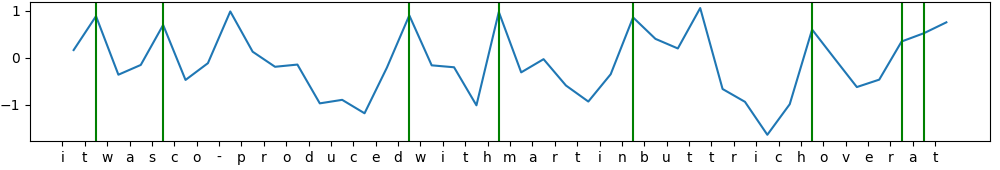}
  \includegraphics[width=0.9\textwidth]{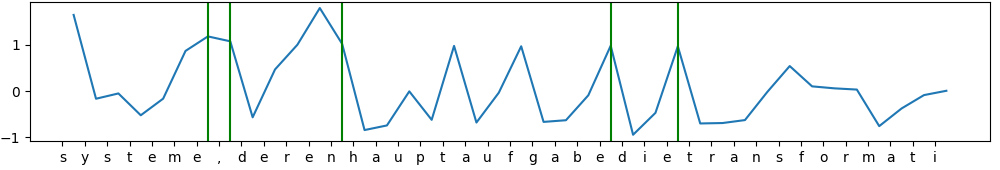}
  \includegraphics[width=0.9\textwidth]{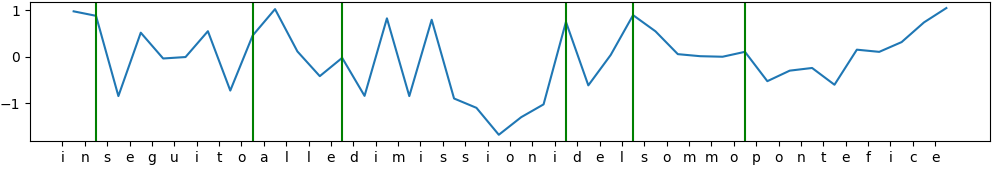}
  \end{center}
  \caption{Examples of the LSTM CNLM boundary unit activation profile, with ground-truth word boundaries marked in
    green. English: \emph{It was co-produced with Martin Buttrich over at\ldots}. German: \emph{Systeme, deren Hauptaufgabe die transformati(-on)} `systems, whose main task is the transformation\ldots'. Italian: \emph{in seguito alle dimissioni del Sommo Pontefice} `following the resignation of the Supreme Pontiff\ldots'. %\textbf{Is there a need to clean up the English and Italian
%      plots? You mentioned adding manual word boundaries?}
	}\label{fig:word-unit}
\end{figure*}

\paragraph{Examples} We looked at the behaviour of the selected LSTM units qualitatively by
extracting random sets of 40-character strings from the development
partition of each language (left-aligned with word onsets) and
plotting the corresponding boundary unit activations. Figure
\ref{fig:word-unit} reports illustrative examples.  In all languages,
most peaks in activation mark word boundaries. However, other
interesting patterns emerge. In English, we see how the unit
reasonably treats \emph{co-} and \emph{produced} in \emph{co-produced}
as separate elements, and it also posits a weaker boundary after the
prefix \emph{pro-}. As it proceeds
left-to-right, with no information on what follows, the network posits a boundary after \emph{but} in \emph{Buttrich}. In
the German example, we observe how the complex word
\emph{Hauptaufgabe} (`main task') is segmented into the morphemes \emph{haupt},
\emph{auf} and \emph{gabe}. Similarly, in the final
\emph{transformati-} fragment, we observe a weak boundary after the prefix
\emph{trans}. In the pronoun \emph{deren} `whose', the case suffix -\emph{n} is separated.
In Italian, \emph{in seguito a} is a lexicalized
multi-word sequence meaning `following' (literally: `in
continuation to'). The boundary unit does not spike inside
it. Similarly, the fixed expression \emph{Sommo Pontefice} (referring
to the Pope) does not trigger inner boundary unit activation spikes.  On the
other hand, we notice peaks after \emph{di} and \emph{mi} in
\emph{dimissioni}. Again, in left-to-right processing, the unit has a
tendency to immediately posit boundaries when frequent function words
are encountered.

% English /home/user/CS_SCR/FAIR18/CHECKPOINTS/boundary-neuron//
% Italian /home/user/CS_SCR/FAIR18/CHECKPOINTS/boundary-neuron//
% German /home/user/CS_SCR/FAIR18/CHECKPOINTS/boundary-neuron//

\paragraph{Detecting word boundaries} To gain a more quantitative
understanding of how well the boundary unit is tracking word
boundaries, we trained a single-parameter diagnostic classifier on the
activation of the unit (the classifier simply sets an optimal
threshold on the unit activation to separate word boundaries from
word-internal positions). We ran two experiments. In the first,
following standard practice, we trained and tested the classifier on
uncontrolled running text. We used 1k characters for training, 1M for
testing, both taken from the left-out Wikipedia test partitions. We
will report F1 performance on this task.

We also considered a more cogent evaluation regime, in which we split
training and test data so that the number of boundary and non-boundary
conditions are balanced, and there is no overlap between training and
test words. Specifically, we randomly selected positions from the test
partitions of the Wikipedia corpus, such that half of these were the
last character of a token, and the other half were not.  % Both
We sampled the test data points subject to the constraint that the
word (in the case of a boundary position) or word prefix (in the case
of a word-internal position) ending at the selected character does not
overlap with the training set. This ensures that a classifier cannot
succeed by looking for encodings reflecting specific words.  For each
datapoint, we fed a substring of the 40 preceding characters to the
CNLM. %
We collected 1,000 such points for training, and tested on 1M additional datapoints. In this
case, we will report classification accuracy as figure of merit.
For reference, in both experiments we also trained diagnostic
classifiers on the \emph{full} hidden layer of the LSTMs.

Looking at the F1 results on uncontrolled running text (Table
\ref{tab:segmentation-results-real}), we observe first that the
LSTM-based full-hidden-layer classifier has strong performance in all
3 languages, confirming that the LSTM model encodes boundary
information.  Moreover, in all languages, a large proportion of this
performance is already accounted for by the single-parameter
classifier using boundary unit activations. This confirms that
tracking boundaries is important enough for the network to devote a
specialized unit to this task. Full-layer RNN results are below
LSTM-level but still strong. There is however a
stronger drop from full-layer to single-unit classification. This is in
line with the fact that, as reported above, the candidate RNN boundary units have lower
boundary correlations than the LSTM ones.

Results for the balanced classifiers tested on new-word
generalization are shown in Table~\ref{tab:segmentation-results}
(because of the different nature of the experiments, these
are not directly comparable to the F1 results in Table~\ref{tab:segmentation-results-real}). Again, we
observe a strong performance of the LSTM-based full-hidden-layer
classifier across the board. The LSTM
single-parameter classifier using boundary unit activations is also strong,  even outperforming the full classifier in German. %
Moreover, in this more cogent setup, the single-unit LSTM classifier
is at least competitive with the full-layer RNN classifier in all languages. The weaker
results of RNNs in the word-centric tasks of the previous sections
might in part be due to their poorer overall ability to track word boundaries,
as specifically suggested by this stricter evaluation setup. %

\begin{table}[t]
	\small
  \begin{center}
    \begin{tabular}{l|l|l|l|l|}
      \multicolumn{1}{c|}{}&\emph{LSTM}&\emph{LSTM}&\emph{RNN}&\emph{RNN}\\
            \multicolumn{1}{c|}{}&\emph{single}&\emph{full}&\emph{single}&\emph{full}\\
      \hline
	    English & 87.7 & 93.0 &  65.6  & 90.5 \\ 
	    German  & 86.6 & 91.9 &  70.4  & 85.0 \\ 
	    Italian & 85.6 & 92.2 &  71.3  & 91.5 \\ 
    \end{tabular}
  \end{center}
  \caption{\label{tab:segmentation-results-real} F1 of single-unit and full-hidden-state word-boundary diagnostic classifiers, trained and tested on uncontrolled running text.}
\end{table}

\begin{table}[t]
	\small
  \begin{center}
    \begin{tabular}{l|l|l|l|l|}
      \multicolumn{1}{c|}{}&\emph{LSTM}&\emph{LSTM}&\emph{RNN}&\emph{RNN}\\
            \multicolumn{1}{c|}{}&\emph{single}&\emph{full}&\emph{single}&\emph{full}\\
      \hline
      English & 77.5& 90.0 & 65.9  & 76.8\\ 
      German  & 80.8& 79.7 & 67.0  & 75.8\\ 
      Italian & 75.5& 82.9 & 71.4  & 75.9\\ 
    \end{tabular}
  \end{center}
  \caption{\label{tab:segmentation-results} Accuracy of  single-unit and full-hidden-state word-boundary diagnostic classifiers, trained and tested on balanced data requiring new-word generalization. Chance accuracy is at 50\%.}
\end{table}

% # python detectBoundariesUnit_Hidden_ExtractPattern_NoWhitespace_Classifier_RealText.py --language english  --batchSize 128 --char_dropout_prob 0.001 --char_embedding_size 200 --char_noise_prob 0.0 --hidden_dim 1024 --language english --layer_num 3 --learning_rate 3.6  --myID 282506230 --load-from wiki-english-nospaces-bptt-282506230 --weight_dropout_hidden 0.01 --weight_dropout_in 0.0

% python detectBoundariesUnit_Hidden_ExtractPattern_NoWhitespace_Classifier_RealText_FullClassifier.py --batchSize 128 --char_dropout_prob 0.0 --char_embedding_size 200 --char_noise_prob 0.0 --hidden_dim 1024 --language italian --layer_num 2 --learning_rate 3.5  --weight_dropout_hidden 0.05 --weight_dropout_in 0.0 --load-from wiki-italian-nospaces-bptt-855947412

\paragraph{Error analysis} As a final way to characterize the function
and behaviour of the boundary units, we inspected the most frequent
under- and over-segmentation errors made by the classifier based on
the single boundary units, in the more difficult balanced task. We
discuss German here, as it is the language where the classifier
reaches highest accuracy, and its tendency to have long,
morphologically complex words makes it particularly
interesting. However, similar patterns were also detected in Italian
and, to a lesser extent, English (in the latter, there are fewer and
less interpretable common oversegmentations, probably because words
are on average shorter and morphology more limited).

Considering first the 30 most common undersegmentations, the large
majority (24/30) are common sequences of grammatical terms or very
frequent items that can sometimes be reasonably re-analyzed as
single function words or adverbs (e.g., \emph{bis zu}, `up to'
(lit.~`until to'), \emph{je nach} `depending on' (lit.~`per after'),
\emph{bis heute} `to date' (lit.~`until today'). 3 cases are
multi-word city names (\emph{Los Angeles}). The final 3 cases
interestingly involve \emph{Bau} `building' followed by \emph{von}
`of' or genitive determiners \emph{der/des}. In its eventive reading,
this noun requires a patient licensed by either a preposition or the
genitive determiner, e.g., \emph{Bau der Mauer} `building of the wall'
(lit.~`building the-GEN wall'). Apparently the model decided to absorb
the case assigner into the form of the noun.

We looked next at the 30 most common oversegmentations, that is, at the
 substrings that were wrongly segmented out of the largest
number of distinct words. We limited the analysis to those containing
at least 3 characters, because shorter strings were ambiguous and
hard to interpret. Among then top oversegmentations, 6 are prefixes that can also occur in
isolation as prepositions or verb particles (\emph{auf} `on',
\emph{nach} `after', etc.). 7 are content words that form many
compounds (e.g., \emph{haupt} `main', occurring in \emph{Hauptstadt}
`capital', \emph{Hauptbahnhof} `main station' etc.; \emph{Land}
`land', occurring in \emph{Deutschland} `Germany', \emph{Landkreis}
`district', etc.). Another 7 items can be classified as suffixes
(e.g., \emph{-lich} as in \emph{s\"udlich} `southern',
\emph{wissenschaftlich} `scientific'), although their segmentation is
not always canonical (e.g., \emph{-chaft} instead of the expected
\emph{-schaft} in \emph{Wissenschaft} `science'). 4 very common
function words are often wrongly segmented out of longer words (e.g.,
\emph{sie} `she' from \emph{sieben} `seven'). The \emph{kom} and
\emph{kon} cases are interesting, as the model
 segments them as stems (or stem fragments) in forms of the verbs \emph{kommen} `to
come' and \emph{k{\"o}nnen} `to be able to', respectively (e.g., \emph{kommt}
and \emph{konnte}), but it also treats them as pseudo-affixes elsewhere
(\emph{komponist} `composer', \emph{kontakt} `contact'). The remaining
3 oversegmentations, \emph{rie}, \emph{run} and \emph{ter} don't have
any clear interpretation.

To conclude, the boundary unit, even when
analyzed through the lens of a classifier that was optimized on
word-level segmentation, is actually tracking salient linguistic
boundaries at different levels. While in many cases these boundaries
naturally coincide with words (hence the high classifier performance),
the CNLM is also sensitive to frequent morphemes and compound
elements, as well as to different types of multi-word
expressions. This is in line with a view of wordhood as a useful but
``soft'', emergent property, rather than a rigid primitive of
linguistic processing.

% python detectBoundariesUnit_Hidden_ExtractPattern_NoWhitespace_Classifier.py --language english  --batchSize 128 --char_dropout_prob 0.001 --char_embedding_size 200 --char_noise_prob 0.0 --hidden_dim 1024 --language english --layer_num 3 --learning_rate 3.6  --myID 282506230 --load-from wiki-english-nospaces-bptt-282506230 --weight_dropout_hidden 0.01 --weight_dropout_in 0.0
% output in 
% results/segmentation-english-frequent-errors-unit-disjoint.txt

\section{Discussion}
\label{sec:discussion}

We probed the linguistic information induced by a character-level LSTM
language model trained on unsegmented text. The model was found to
possess implicit knowledge about a range of intuitively word-mediated
phenomena, such as sensitivity to lexical categories and syntactic and
shallow-semantics dependencies. A model initialized with a word
vocabulary and fed tokenized input was in general superior, but the
performance of the word-less model did not lag much behind, suggesting
that word priors are helpful but not strictly required. A
character-level RNN was less consistent than the LSTM, suggesting that
the latter's ability to track information across longer time spans is
important to make the correct generalizations. The character-level
models consistently outperformed n-gram controls, confirming they are
tapping into more abstract patterns than local co-occurrence
statistics.

As a first step towards understanding \emph{how} character-level
models handle supra-character phenomena, we searched and found
specialized boundary-tracking units in them. These units are not only
and not always sensitive to word boundaries, but also respond to other
salient items, such as morphemes and multi-word expressions, in
accordance with an ``emergent'' and flexible view of the basic
constituents of language \cite{Schiering:etal:2010}.

Our results are preliminary in many ways. Our tests are relatively
simple. We did not attempt, for example, to model long-distance
agreement in presence of distractors, a challenging task even for
humans \citep{Gulordava:etal:2018}. The results on number
classification in German suggest that the models might not be
capturing linguistic generalizations of the correct degree of
abstractness, settling for shallower heuristics. Still, as a whole,
our work suggests that a large corpus, combined with the weak priors
encoded in an LSTM, might suffice to learn generalizations about
word-mediated linguistic processes without a hard-coded word lexicon
or explicit wordhood cues.

Nearly all contemporary linguistics recognizes a central role to the
lexicon \cite[see, e.g.,][for very different
perspectives]{Sag:etal:2003,Goldberg:2005,Radford:2006,Bresnan:etal:2016,Jezek:2016}. Linguistic
formalisms assume that the lexicon is essentially a dictionary of
words, possibly complemented by other units, not unlike the list of
words and associated embeddings in a standard word-based
NLM. Intriguingly, our CNLMs captured a range of lexical phenomena
\emph{without} anything resembling a word dictionary. Any information
a CNLM might acquire about units larger than characters must be stored
in its recurrent weights. This suggests a radically different and
possibly more neurally plausible view of the lexicon as implicitly
encoded in a distributed memory, that we intend to characterize more
precisely and test in future work \cite[similar ideas are being
explored in a more applied NLP perspective,
e.g.,][]{gillick2016multilingual,lee2017fully,cherry2018revisiting}.

Concerning the model input, we would like to study whether the
CNLM successes crucially depend on the huge amount
of training data it receives.  Are word priors more important when
learning from smaller corpora? In terms of comparison with human
learning, the Wikipedia text we fed our CNLMs is far from
what children acquiring a language would hear. Future work should
explore character/phoneme-level learning from child-directed speech
corpora. Still, by feeding our networks ``grown-up'' prose, we are
arguably making the job of identifying basic constituents harder than
it might be when processing the simpler utterances of early
child-directed speech \cite{Tomasello:2003}.

As discussed, a rigid word notion is problematic both
cross-linguistically (cf.~polysynthetic and agglutinative languages)
and within single linguistic systems \cite[cf.~the view that
the lexicon hosts units at different levels of the linguistic
hierarchy, from morphemes to large syntactic constructions,
e.g.,][]{Jackendoff:1997,Croft:Cruse:2004,Goldberg:2005}. This study provided a necessary initial check
that word-free models can account for phenomena traditionally
seen as word-based. Future work should test whether such models
 can also account for grammatical patterns that are harder to
capture in word-based formalisms, exploring both a typologically
wider range of languages and a broader set of grammatical tests.

\section*{Acknowledgments}

We would like to thank Piotr Bojanowski, Alex Cristia, Kristina
Gulordava, Urvashi Khandelwal, Germ\'{a}n Kruszewski, Sebastian
Riedel, Hinrich Sch\"{u}tze and the anonymous reviewers for feedback
and advice.

\bibliography{marco,michael}

\begin{thebibliography}{77}
\expandafter\ifx\csname natexlab\endcsname\relax\def\natexlab#1{#1}\fi

\bibitem[{Adi et~al.(2017)Adi, Kermany, Belinkov, Lavi, and
  Goldberg}]{Adi:etal:2017}
Yossi Adi, Einat Kermany, Yonatan Belinkov, Ofer Lavi, and Yoav Goldberg. 2017.
\newblock Fine-grained analysis of sentence embeddings using auxiliary
  prediction tasks.
\newblock In \emph{Proceedings of ICLR Conference Track}, Toulon, France.
\newblock Published online:
  \url{https://openreview.net/group?id=ICLR.cc/2017/conference}.

\bibitem[{Alishahi et~al.(2017)Alishahi, Barking, and
  Chrupa{\l}a}]{Alishahi:etal:2017}
Afra Alishahi, Marie Barking, and Grzegorz Chrupa{\l}a. 2017.
\newblock Encoding of phonology in a recurrent neural model of grounded speech.
\newblock In \emph{Proceedings of CoNLL}, pages 368--378, Vancouver, Canada.

\bibitem[{Bar(2007)}]{Bar:2007}
Moshe Bar. 2007.
\newblock The proactive brain: using analogies and associations to generate
  predictions.
\newblock \emph{Trends in Cognitive Science}, 11(7):280--289.

\bibitem[{Belinkov et~al.(2017)Belinkov, Durrani, Dalvi, Sajjad, and
  Glass}]{Belinkov:etal:2017}
Yonatan Belinkov, Nadir Durrani, Fahim Dalvi, Hassan Sajjad, and James Glass.
  2017.
\newblock What do neural machine translation models learn about morphology?
\newblock In \emph{Proceedings of ACL}, pages 861--872, Vancouver, Canada.

\bibitem[{Bickel and {Z\'{u}\~{n}iga}(2017)}]{Bickel:Zuniga:2017}
Balthasar Bickel and Fernando {Z\'{u}\~{n}iga}. 2017.
\newblock The `word' in polysynthetic languages: Phonological and syntactic
  challenges.
\newblock In Michael Fortescue, Marianne Mithun, and Nicholas Evans, editors,
  \emph{Oxford Handbook of Polysynthesis}, pages 158--186. Oxford University
  Press, Oxford, UK.

\bibitem[{Bojanowski et~al.(2016)Bojanowski, Joulin, and
  Mikolov}]{Bojanowski:etal:2016}
Piotr Bojanowski, Armand Joulin, and Tomas Mikolov. 2016.
\newblock Alternative structures for character-level {RNNs}.
\newblock In \emph{Proceedings of ICLR Workshop Track}, San Juan, Puerto Rico.
\newblock Published online:
  \url{https://openreview.net/group?id=ICLR.cc/2016/workshop}.

\bibitem[{Brants et~al.(2002)Brants, Dipper, Hansen, Lezius, and
  Smith}]{brants2002tiger}
Sabine Brants, Stefanie Dipper, Silvia Hansen, Wolfgang Lezius, and George
  Smith. 2002.
\newblock The {TIGER} treebank.
\newblock In \emph{Proceedings of the workshop on treebanks and linguistic
  theories}, volume 168.

\bibitem[{Brent and Cartwright(1996)}]{Brent:Cartwright:1996}
Michael Brent and Timothy Cartwright. 1996.
\newblock Distributional regularity and phonotactic constraints are useful for
  segmentation.
\newblock \emph{Cognition}, 61:93--125.

\bibitem[{Bresnan et~al.(2016)Bresnan, Asudeh, Toivonen, and
  Wechsler}]{Bresnan:etal:2016}
Joan Bresnan, Ash Asudeh, Ida Toivonen, and Stephen Wechsler. 2016.
\newblock \emph{Lexical-Functional Syntax, 2nd ed.}
\newblock Blackwell, Malden, MA.

\bibitem[{Bresnan and Mchombo(1995)}]{Bresnan:Mchombo:1995}
Joan Bresnan and Sam Mchombo. 1995.
\newblock The lexical integrity principle: Evidence from {Bantu}.
\newblock \emph{Natural Language and Linguistic Theory}, pages 181--254.

\bibitem[{Cherry et~al.(2018)Cherry, Foster, Bapna, Firat, and
  Macherey}]{cherry2018revisiting}
Colin Cherry, George Foster, Ankur Bapna, Orhan Firat, and Wolfgang Macherey.
  2018.
\newblock Revisiting character-based neural machine translation with capacity
  and compression.
\newblock \emph{arXiv preprint arXiv:1808.09943}.

\bibitem[{Chomsky(1970)}]{Chomsky:1970}
Noam Chomsky. 1970.
\newblock Remarks on nominalization.
\newblock In Roderick Jacobs and Peter Rosenbaum, editors, \emph{Readings in
  {English} Transformational Grammar}, pages 184--221. Ginn, Waltham, MA.

\bibitem[{Christiansen et~al.(2005)Christiansen, Conway, and
  Curtin}]{Christiansen:etal:2005}
Morten Christiansen, Christopher Conway, and Suzanne Curtin. 2005.
\newblock Multiple-cue integration in language acquisition: A connectionist
  model of speech segmentation and rule-like behavior.
\newblock In James Minett and William Wang, editors, \emph{Language
  Acquisition, Change and Emergence: Essays in Evolutionary Linguistics}, pages
  205--249. City University of Hong Kong Press, Hong Kong.

\bibitem[{Christiansen et~al.(1998)Christiansen, Joseh, and
  Seidenberg}]{Christiansen:etal:1998}
Morten Christiansen, Allen Joseh, and Mark Seidenberg. 1998.
\newblock Learning to segment speech using multiple cues: A connectionist
  model.
\newblock \emph{Language and Cognitive Processes}, 13(2/3):221--268.

\bibitem[{Clark(2016)}]{Clark:2016}
Andy Clark. 2016.
\newblock \emph{Surfing Uncertainty}.
\newblock Oxford University Press, Oxford, UK.

\bibitem[{Conneau et~al.(2018)Conneau, Kruszewski, Lample, Barrault, and
  Baroni}]{Conneau:etal:2018}
Alexis Conneau, Germ{\'a}n Kruszewski, Guillaume Lample, Lo{\"i}c Barrault, and
  Marco Baroni. 2018.
\newblock What you can cram into a single {\$}{\&}!{\#}* vector: Probing
  sentence embeddings for linguistic properties.
\newblock In \emph{Proceedings ACL}, pages 2126--2136, Melbourne, Australia.

\bibitem[{Cotterell et~al.(2018)Cotterell, Mielke, Eisner, and
  Roark}]{cotterell2018all}
Ryan Cotterell, Sebastian~J Mielke, Jason Eisner, and Brian Roark. 2018.
\newblock Are all languages equally hard to language-model?
\newblock In \emph{Proceedings of the 2018 Conference of the North American
  Chapter of the Association for Computational Linguistics: Human Language
  Technologies, Volume 2 (Short Papers)}, volume~2, pages 536--541.

\bibitem[{Croft and Cruse(2004)}]{Croft:Cruse:2004}
William Croft and Alan Cruse. 2004.
\newblock \emph{Cognitive Linguistics}.
\newblock Cambridge University Press, Cambridge, UK.

\bibitem[{{Di Sciullo} and Williams(1987)}]{DiSciullo:Williams:1987}
Anna-Maria {Di Sciullo} and Edwin Williams. 1987.
\newblock \emph{On the Definition of Word}.
\newblock MIT Press, Cambridge, MA.

\bibitem[{Dixon and Aikhenvald(2002)}]{Dixon:Aikhenvald:2002}
Robert Dixon and Alexandra Aikhenvald, editors. 2002.
\newblock \emph{Word: A cross-linguistic typology}.
\newblock Cambridge University Press, Cambridge, UK.

\bibitem[{Dudenredaktion(2019)}]{duden2019mit}
Dudenredaktion. 2019.
\newblock mit ({A}dverb).
\newblock In \emph{Duden online}.
\newblock \url{https://www.duden.de/node/152710/revision/152746}, retrieved
  June 3, 2019.

\bibitem[{Elman(1990)}]{Elman:1990}
Jeffrey Elman. 1990.
\newblock Finding structure in time.
\newblock \emph{Cognitive Science}, 14:179--211.

\bibitem[{Ettinger et~al.(2018)Ettinger, Elgohary, Phillips, and
  Resnik}]{Ettinger:etal:2018}
Allyson Ettinger, Ahmed Elgohary, Colin Phillips, and Philip Resnik. 2018.
\newblock Assessing composition in sentence vector representations.
\newblock In \emph{Proceedings of COLING}, pages 1790--1801, Santa Fe, NM.

\bibitem[{Frank et~al.(2013)Frank, Mathis, and Badecker}]{Frank:etal:2013}
Robert Frank, Donald Mathis, and William Badecker. 2013.
\newblock The acquisition of anaphora by simple recurrent networks.
\newblock \emph{Language Acquisition}, 20(3):181--227.

\bibitem[{Fusi et~al.(2016)Fusi, Miller, and Rigotti}]{Fusi:etal:2016}
Stefano Fusi, Earl Miller, and Mattia Rigotti. 2016.
\newblock Why neurons mix: High dimensionality for higher cognition.
\newblock \emph{Current Opinion in Neurobiology}, 37:66--74.

\bibitem[{Gerz et~al.(2018)Gerz, Vuli\'{c}, Ponti, Naradowsky, Reichart, and
  Korhonen}]{Gerz:etal:2018}
Daniela Gerz, Ivan Vuli\'{c}, Edoardo~Maria Ponti, Jason Naradowsky, Roi
  Reichart, and Anna Korhonen. 2018.
\newblock Language modeling for morphologically rich languages: Character-aware
  modeling for word-level prediction.
\newblock \emph{Transactions of the Association for Computational Linguistics},
  6:451--465.

\bibitem[{Gillick et~al.(2016)Gillick, Brunk, Vinyals, and
  Subramanya}]{gillick2016multilingual}
Dan Gillick, Cliff Brunk, Oriol Vinyals, and Amarnag Subramanya. 2016.
\newblock Multilingual language processing from bytes.
\newblock In \emph{Proceedings of NAACL-HLT}, pages 1296--1306.

\bibitem[{Godin et~al.(2018)Godin, Demuynck, Dambre, {De Neve}, and
  Demeester}]{Godin:etal:2018}
Fr\'{e}deric Godin, Kris Demuynck, Joni Dambre, Wesley {De Neve}, and Thomas
  Demeester. 2018.
\newblock Explaining character-aware neural networks for word-level prediction:
  Do they discover linguistic rules?
\newblock In \emph{Proceedings of EMNLP}, Brussels, Belgium.
\newblock {I}n press.

\bibitem[{Goldberg(2005)}]{Goldberg:2005}
Adele Goldberg. 2005.
\newblock \emph{Constructions at work: The nature of generalization in
  language}.
\newblock Oxford University Press, Oxford, UK.

\bibitem[{Goldberg(2017)}]{Goldberg:2017}
Yoav Goldberg. 2017.
\newblock \emph{Neural Network Methods for Natural Language Processing}.
\newblock Morgan {\&} Claypool, San Francisco, CA.

\bibitem[{Goldwater et~al.(2009)Goldwater, Griffiths, and
  Johnson}]{goldwater-bayesian-2009}
Sharon Goldwater, Thomas~L. Griffiths, and Mark Johnson. 2009.
\newblock A {Bayesian} framework for word segmentation: {Exploring} the effects
  of context.
\newblock \emph{Cognition}, 112(1):21--54.

\bibitem[{Graves(2014)}]{DBLP:journals/corr/Graves13}
Alex Graves. 2014.
\newblock \href {https://arxiv.org/abs/1308.0850} {Generating sequences with
  recurrent neural networks}.
\newblock \emph{CoRR}, abs/1308.0850v5.

\bibitem[{Gulordava et~al.(2018)Gulordava, Bojanowski, Grave, Linzen, and
  Baroni}]{Gulordava:etal:2018}
Kristina Gulordava, Piotr Bojanowski, Edouard Grave, Tal Linzen, and Marco
  Baroni. 2018.
\newblock Colorless green recurrent networks dream hierarchically.
\newblock In \emph{Proceedings of NAACL}, pages 1195--1205, New Orleans, LA.

\bibitem[{Haspelmath(2011)}]{Haspelmath:2011}
Martin Haspelmath. 2011.
\newblock The indeterminacy of word segmentation and the nature of morphology
  and syntax.
\newblock \emph{Folia Linguistica}, 45(1):31--80.

\bibitem[{Hochreiter and Schmidhuber(1997)}]{Hochreiter:Schmidhuber:1997}
Sepp Hochreiter and J\"{u}rgen Schmidhuber. 1997.
\newblock Long short-term memory.
\newblock \emph{Neural Computation}, 9(8):1735--1780.

\bibitem[{Hupkes et~al.(2018)Hupkes, Veldhoen, and Zuidema}]{Hupkes:etal:2017}
Dieuwke Hupkes, Sara Veldhoen, and Willem Zuidema. 2018.
\newblock Visualisation and `diagnostic classifiers' reveal how recurrent and
  recursive neural networks process hierarchical structure.
\newblock \emph{Journal of Artificial Intelligence Research}, 61:907--926.

\bibitem[{Jackendoff(1997)}]{Jackendoff:1997}
Ray Jackendoff. 1997.
\newblock Twistin' the night away.
\newblock \emph{Language}, 73:534--559.

\bibitem[{Jackendoff(2002)}]{Jackendoff:2002}
Ray Jackendoff. 2002.
\newblock \emph{Foundations of Language: Brain, Meaning, Grammar, Evolution}.
\newblock Oxford University Press, Oxford, UK.

\bibitem[{Je\v{z}ek(2016)}]{Jezek:2016}
Elisabetta Je\v{z}ek. 2016.
\newblock \emph{The Lexicon: An Introduction}.
\newblock Oxford University Press, Oxford, UK.

\bibitem[{K\`ad\`ar et~al.(2017)K\`ad\`ar, Chrupa\l{}a, and
  Alishahi}]{Kadar:etal:2017}
\`Akos K\`ad\`ar, Grzegorz Chrupa\l{}a, and Afra Alishahi. 2017.
\newblock Representation of linguistic form and function in recurrent neural
  networks.
\newblock \emph{Computational Linguistics}, 43(4):761--780.

\bibitem[{Kamper et~al.(2016)Kamper, Jansen, and Goldwater}]{Kamper:etal:2016}
Herman Kamper, Aren Jansen, and Sharon Goldwater. 2016.
\newblock Unsupervised word segmentation and lexicon discovery using acoustic
  word embeddings.
\newblock \emph{IEEE Transactions on Audio, Speech and Language Processing},
  24(4):669--679.

\bibitem[{Kann et~al.(2016)Kann, Cotterell, and Sch{\"u}tze}]{Kann:etal:2016}
Katharina Kann, Ryan Cotterell, and Hinrich Sch{\"u}tze. 2016.
\newblock Neural morphological analysis: Encoding-decoding canonical segments.
\newblock In \emph{Proceedings of EMNLP}, pages 961--967, Austin, Texas.

\bibitem[{Kementchedjhieva and Lopez(2018)}]{Kementchedjhieva:Lopez:2018}
Yova Kementchedjhieva and Adam Lopez. 2018.
\newblock {`Indicatements'} that character language models learn {English}
  morpho-syntactic units and regularities.
\newblock In \emph{Proceedings of the EMNLP BlackboxNLP Workshop}, pages
  145--153, Brussels, Belgium.

\bibitem[{Kim et~al.(2016)Kim, Jernite, Sontag, and Rush}]{Kim:etal:2016}
Yoon Kim, Yacine Jernite, David Sontag, and Alexander Rush. 2016.
\newblock Character-aware neural language models.
\newblock In \emph{Proceedings of AAAI}, pages 2741--2749, Phoenix, AZ.

\bibitem[{Kirov and Cotterell(2018)}]{Kirov:Cotterell:2018}
Christo Kirov and Ryan Cotterell. 2018.
\newblock Recurrent neural networks in linguistic theory: Revisiting {Pinker
  and Prince} (1988) and the past tense debate.
\newblock \emph{Transactions of the Association for Computational Linguistics}.
\newblock {I}n press.

\bibitem[{Kuhl(2004)}]{Kuhl:2004}
Patricia Kuhl. 2004.
\newblock Early language acquisition: Cracking the speech code.
\newblock \emph{Nature Reviews Neuroscience}, 5(11):831--843.

\bibitem[{Lau et~al.(2017)Lau, Clark, and Lappin}]{Lau:etal:2017}
Jey~Han Lau, Alexander Clark, and Shalom Lappin. 2017.
\newblock Grammaticality, acceptability, and probability: A probabilistic view
  of linguistic knowledge.
\newblock \emph{Cognitive Science}, 41(5):1202--1241.

\bibitem[{Lee et~al.(2017)Lee, Cho, and Hofmann}]{lee2017fully}
Jason Lee, Kyunghyun Cho, and Thomas Hofmann. 2017.
\newblock Fully character-level neural machine translation without explicit
  segmentation.
\newblock \emph{Transactions of the Association for Computational Linguistics},
  5:365--378.

\bibitem[{Li et~al.(2016)Li, Chen, Hovy, and Jurafsky}]{Li:etal:2016}
Jiwei Li, Xinlei Chen, Eduard Hovy, and Dan Jurafsky. 2016.
\newblock Visualizing and understanding neural models in {NLP}.
\newblock In \emph{Proceedings of NAACL}, pages 681--691, San Diego, CA.

\bibitem[{Linzen et~al.(2018)Linzen, Chrupa{\l}a, and
  Alishahi}]{Linzen:etal:2018}
Tal Linzen, Grzegorz Chrupa{\l}a, and Afra Alishahi, editors. 2018.
\newblock \emph{Proceedings of the EMNLP BlackboxNLP Workshop}.
\newblock ACL, Brussels, Belgium.

\bibitem[{Linzen et~al.(2016)Linzen, Dupoux, and Goldberg}]{Linzen:etal:2016}
Tal Linzen, Emmanuel Dupoux, and Yoav Goldberg. 2016.
\newblock Assessing the ability of {LSTM}s to learn syntax-sensitive
  dependencies.
\newblock \emph{Transactions of the Association for Computational Linguistics},
  4:521--535.

\bibitem[{Maye et~al.(2002)Maye, Werker, and Gerken}]{Maye:etal:2002}
Jessica Maye, Janet Werker, and LouAnn Gerken. 2002.
\newblock Infant sensitivity to distributional information can affect phonetic
  discrimination.
\newblock \emph{Cognition}, 82(3):B101--B111.

\bibitem[{McCoy et~al.(2018)McCoy, Frank, and Linzen}]{McCoy:etal:2018}
Thomas McCoy, Robert Frank, and Tal Linzen. 2018.
\newblock Revisiting the poverty of the stimulus: Hierarchical generalization
  without a hierarchical bias in recurrent neural networks.
\newblock In \emph{Proceedings of CogSci}, pages 2093--2098, Madison, WI.

\bibitem[{McDonald et~al.(2013)McDonald, Nivre, Quirmbach-Brundage, Goldberg,
  Das, Ganchev, Hall, Petrov, Zhang, T{\"a}ckstr{\"o}m
  et~al.}]{mcdonald2013universal}
Ryan McDonald, Joakim Nivre, Yvonne Quirmbach-Brundage, Yoav Goldberg, Dipanjan
  Das, Kuzman Ganchev, Keith Hall, Slav Petrov, Hao Zhang, Oscar
  T{\"a}ckstr{\"o}m, et~al. 2013.
\newblock Universal dependency annotation for multilingual parsing.
\newblock In \emph{Proceedings of the 51st Annual Meeting of the Association
  for Computational Linguistics (Volume 2: Short Papers)}, volume~2, pages
  92--97.

\bibitem[{Melamud et~al.(2016)Melamud, Goldberger, and
  Dagan}]{melamud2016context2vec}
Oren Melamud, Jacob Goldberger, and Ido Dagan. 2016.
\newblock context2vec: Learning generic context embedding with bidirectional
  lstm.
\newblock In \emph{Proceedings of The 20th SIGNLL Conference on Computational
  Natural Language Learning}, pages 51--61.

\bibitem[{Merity et~al.(2018)Merity, Keskar, and Socher}]{merity2018analysis}
Stephen Merity, Nitish~Shirish Keskar, and Richard Socher. 2018.
\newblock An analysis of neural language modeling at multiple scales.
\newblock \emph{arXiv preprint arXiv:1803.08240}.

\bibitem[{Mikolov(2012)}]{Mikolov:2012}
Tomas Mikolov. 2012.
\newblock \emph{Statistical language models based on neural networks}.
\newblock Dissertation, Brno University of Technology.

\bibitem[{Mikolov et~al.(2013{\natexlab{a}})Mikolov, Chen, Corrado, and
  Dean}]{DBLP:journals/corr/abs-1301-3781}
Tomas Mikolov, Kai Chen, Greg Corrado, and Jeffrey Dean. 2013{\natexlab{a}}.
\newblock \href {http://arxiv.org/abs/1301.3781} {Efficient estimation of word
  representations in vector space}.
\newblock \emph{CoRR}, abs/1301.3781.

\bibitem[{Mikolov et~al.(2011)Mikolov, Sutskever, Deoras, Le, Kombrink, and
  Cernock{\'{y}}}]{Mikolov:etal:2011}
Tomas Mikolov, Ilya Sutskever, Anoop Deoras, Hai-Son Le, Stefan Kombrink, and
  Jan Cernock{\'{y}}. 2011.
\newblock Subword language modeling with neural networks.
\newblock \url{http://www.fit.vutbr.cz/~imikolov/rnnlm/}.

\bibitem[{Mikolov et~al.(2013{\natexlab{b}})Mikolov, Yih, and
  Zweig}]{Mikolov:etal:2013a}
Tomas Mikolov, Wen-tau Yih, and Geoffrey Zweig. 2013{\natexlab{b}}.
\newblock Linguistic regularities in continuous space word representations.
\newblock In \emph{Proceedings of NAACL}, pages 746--751, Atlanta, Georgia.

\bibitem[{Pater(2018)}]{Pater:2018}
Joe Pater. 2018.
\newblock Generative linguistics and neural networks at 60: Foundation,
  friction, and fusion.
\newblock \emph{Language}.
\newblock {I}n press.

\bibitem[{Radford et~al.(2017)Radford, J{\'{o}}zefowicz, and
  Sutskever}]{DBLP:journals/corr/RadfordJS17}
Alec Radford, Rafal J{\'{o}}zefowicz, and Ilya Sutskever. 2017.
\newblock \href {http://arxiv.org/abs/1704.01444} {Learning to generate reviews
  and discovering sentiment}.
\newblock \emph{CoRR}, abs/1704.01444.

\bibitem[{Radford(2006)}]{Radford:2006}
Andrew Radford. 2006.
\newblock Minimalist syntax revisited.
\newblock \url{http://www.public.asu.edu/~gelderen/Radford2009.pdf}.

\bibitem[{Sag et~al.(2003)Sag, Wasow, and Bender}]{Sag:etal:2003}
Ivan Sag, Thomas Wasow, and Emily Bender. 2003.
\newblock \emph{Syntactic Theory: A Formal Introduction}.
\newblock CSLI, Stanford, CA.

\bibitem[{Schiering et~al.(2010)Schiering, Bickel, and
  Hildebrandt}]{Schiering:etal:2010}
Ren\'{e} Schiering, Balthasar Bickel, and Kristine Hildebrandt. 2010.
\newblock The prosodic word is not universal, but emergent.
\newblock \emph{Journal of Linguistics}, 46(3):657--709.

\bibitem[{Schmid(1999)}]{schmid1999improvements}
Helmut Schmid. 1999.
\newblock Improvements in part-of-speech tagging with an application to german.
\newblock In \emph{Natural language processing using very large corpora}, pages
  13--25. Springer.

\bibitem[{Sch{\"u}tze(2017)}]{Schuetze:2017}
Hinrich Sch{\"u}tze. 2017.
\newblock Nonsymbolic text representation.
\newblock In \emph{Proceedings of EACL}, pages 785--796, Valencia, Spain.

\bibitem[{Sennrich(2017)}]{Sennrich:2017}
Rico Sennrich. 2017.
\newblock How grammatical is character-level neural machine translation?
  assessing {MT} quality with contrastive translation pairs.
\newblock In \emph{Proceedings of EACL (Short Papers)}, pages 376--382,
  Valencia, Spain.

\bibitem[{Shi et~al.(2016)Shi, Padhi, and Knight}]{Shi:etal:2016}
Xing Shi, Inkit Padhi, and Kevin Knight. 2016.
\newblock Does string-based neural {MT} learn source syntax?
\newblock In \emph{Proceedings of EMNLP}, pages 1526--1534, Austin, Texas.

\bibitem[{Sutskever et~al.(2011)Sutskever, Martens, and
  Hinton}]{Sutskever:etal:2011}
Ilya Sutskever, James Martens, and Geoffrey Hinton. 2011.
\newblock Generating text with recurrent neural networks.
\newblock In \emph{Proceedings of ICML}, pages 1017--1024, Bellevue, WA.

\bibitem[{Sutskever et~al.(2014)Sutskever, Vinyals, and
  Le}]{sutskever2014sequence}
Ilya Sutskever, Oriol Vinyals, and Quoc~V Le. 2014.
\newblock Sequence to sequence learning with neural networks.
\newblock In \emph{Advances in neural information processing systems}, pages
  3104--3112.

\bibitem[{Tomasello(2003)}]{Tomasello:2003}
Michael Tomasello. 2003.
\newblock \emph{Constructing a Language: A Usage-Based Theory of Language
  Acquisition}.
\newblock Harvard University Press, Cambridge, MA.

\bibitem[{Williams(2007)}]{Williams:2007}
Edwin Williams. 2007.
\newblock Dumping lexicalism.
\newblock In Gillian Ramchand and Charles Reiss, editors, \emph{The Oxford
  Handbook of Linguistic Interfaces}. Oxford University Press, Oxford, UK.

\bibitem[{Woods(2016)}]{woods2016exploiting}
Aubrie Woods. 2016.
\newblock Exploiting linguistic features for sentence completion.
\newblock In \emph{Proceedings of the 54th Annual Meeting of the Association
  for Computational Linguistics (Volume 2: Short Papers)}, volume~2, pages
  438--442.

\bibitem[{Zhang et~al.(2016)Zhang, Lu, and Lapata}]{zhang2016top}
Xingxing Zhang, Liang Lu, and Mirella Lapata. 2016.
\newblock Top-down tree long short-term memory networks.
\newblock In \emph{Proceedings of the 2016 Conference of the North American
  Chapter of the Association for Computational Linguistics: Human Language
  Technologies}, pages 310--320.

\bibitem[{Zweig and Burges(2011)}]{Zweig:Burges:2011}
Geoffrey Zweig and Christopher Burges. 2011.
\newblock The {Microsoft Research} sentence completion challenge.
\newblock Technical Report MSR-TR-2011-129, Microsoft Research.

\bibitem[{Zweig et~al.(2012)Zweig, Platt, Meek, Burges, Yessenalina, and
  Liu}]{zweig2012computational}
Geoffrey Zweig, John~C Platt, Christopher Meek, Christopher~JC Burges, Ainur
  Yessenalina, and Qiang Liu. 2012.
\newblock Computational approaches to sentence completion.
\newblock In \emph{Proceedings of the 50th Annual Meeting of the Association
  for Computational Linguistics: Long Papers-Volume 1}, pages 601--610.
  Association for Computational Linguistics.

\end{thebibliography}
\bibliographystyle{acl_natbib}

 \appendix

 \begin{table*}[t]
 	\begin{tabular}{l|lll|lll|lllllll}
 		&  \multicolumn{3}{c}{LSTM} & \multicolumn{3}{|c|}{RNN} & \multicolumn{3}{c}{WordNLM} \\
 		       &  En.     &  Ge.    & It.    & En.    &    Ge.   &  It.     &  En.     &   Ge.   &    It. \\  \hline
 	Batch Size     &  128   &  512  & 128  & 256  & 256    &  256   &  128   &   128 &  128   \\              
 	Embedding Size &  200   &  100  & 200  & 200  & 50     &  50    &  1024  &   200 &  200   \\             
 	Dimension      &  1024  &  1024 & 1024 & 2048 & 2048   &  2048  &  1024  &  1024 &  1024  \\  
 	Layers         &  3     &  2    & 2    & 2    & 2      &  2     &  2     &  2    &  2     \\   
 	Learning Rate  &  3.6   &  2.0  & 3.2  & 0.01 & 0.1    &  0.1   &  1.1   &  0.9  &  1.2   \\ 
 	Decay          &  0.95  &  1.0  & 0.98 & 0.9  & 0.95   &  0.95  &  1.0   &  1.0  &  0.98  \\
 	BPTT Length    &  80    &  50   & 80   & 50   & 30     &  30    &  50    &  50   &  50    \\
 	Hidden Dropout &  0.01  &  0.0  & 0.0  & 0.05 & 0.0    &  0.0   &  0.15  &  0.15 &  0.05  \\   
 	Embedding Dropout  & 0.0& 0.01  & 0.0  & 0.01 & 0.0    &  0.0   &  0.0   &  0.1  &  0.0   \\   
 	Input Dropout  & 0.001 &  0.0   & 0.0  & 0.001& 0.01   &  0.01  &  0.01  &  0.001&  0.01  \\ 
         Nonlinearity   &   --  & --     & --   & ReLu & tanh   &  tanh  &   --   &  --   &  --    \\                   
 \end{tabular}
 	\caption{Chosen hyperparameters}
 \end{table*}

\end{document}